\documentclass[11pt,a4paper,table]{article}

\usepackage[]{coling2020}

\usepackage[a-1b]{pdfx}   % for PDF/A-1b

\hypersetup{colorlinks=true,linkcolor=black,citecolor=black,filecolor=black,urlcolor=black}

\usepackage{natbib}
\renewcommand{\newcite}[1]{\citet{#1}}
\renewcommand{\cite}[1]{\citep{#1}}

\usepackage{times}
\usepackage{latexsym}
\usepackage{url}
\usepackage{wrapfig}

\makeatletter
\newcommand*{\addFileDependency}[1]{% argument=file name and extension
  \typeout{(#1)}
  \@addtofilelist{#1}
  \IfFileExists{#1}{}{\typeout{No file #1.}}
}
\makeatother

\usepackage[T1]{fontenc}
\usepackage[scaled=.8]{beramono}
\usepackage[scaled=.95]{helvet}

\usepackage{amsmath}
\usepackage{color,soul}

\usepackage{subcaption}

\usepackage[shortlabels]{enumitem} % customizable lists

\usepackage{multirow}
\usepackage{tikz}
\usepackage{tikz-dependency}
\usepackage[warn]{textcomp}
\usepackage{multirow}
\usepackage{etoolbox}
\usepackage{xr}
\usepackage{xfrac}
\usepackage{pgfplotstable}
\usepackage{amsmath}
\usepackage{amssymb}
\usepackage{tabularx}
\usepackage{collcell}
\usepackage{booktabs}
\usepackage{pifont}
\usepackage{nameref}
\usepackage{cleveref}

\pgfplotsset{compat=1.15}
\crefformat{part}{\S#2#1#3}
\crefformat{chapter}{\S#2#1#3}
\crefformat{section}{\S#2#1#3}
\crefformat{subsection}{\S#2#1#3}
\crefformat{subsubsection}{\S#2#1#3}
\crefformat{paragraph}{\P#2#1#3}
\crefformat{subparagraph}{\P#2#1#3}
\crefmultiformat{section}{\S#2#1#3}{ and~\S#2#1#3}{, \S#2#1#3}{, and~\S#2#1#3}
\crefmultiformat{subsection}{\S#2#1#3}{ and~\S#2#1#3}{, \S#2#1#3}{, and~\S#2#1#3}
\crefmultiformat{subsubsection}{\S#2#1#3}{ and~\S#2#1#3}{, \S#2#1#3}{, and~\S#2#1#3}
\crefmultiformat{paragraph}{\P\P#2#1#3}{ and~#2#1#3}{, #2#1#3}{, and~#2#1#3}
\crefmultiformat{subparagraph}{\P\P#2#1#3}{ and~#2#1#3}{, #2#1#3}{, and~#2#1#3}
\crefrangeformat{section}{\mbox{\S\S#3#1#4--#5#2#6}}
\crefrangeformat{subsection}{\mbox{\S\S#3#1#4--#5#2#6}}
\crefrangeformat{subsubsection}{\mbox{\S\S#3#1#4--#5#2#6}}
\crefrangeformat{paragraph}{\mbox{\P\P#3#1#4--#5#2#6}}
\crefrangeformat{subparagraph}{\mbox{\P\P#3#1#4--#5#2#6}}
\crefname{part}{Part}{Parts}
\Crefname{part}{Part}{Parts}
\crefname{chapter}{ch.}{ch.}
\Crefname{chapter}{Ch.}{Ch.}
\crefname{figure}{figure}{figures}
\crefname{subfigure}{figure}{figures}
\Crefname{subfigure}{Figure}{Figures}
\crefname{appsec}{appendix}{appendices}
\Crefname{appsec}{Appendix}{Appendices}
\crefname{algocf}{algorithm}{algorithms}
\Crefname{algocf}{Algorithm}{Algorithms}
\crefname{enums,enumsi}{example}{examples}
\Crefname{enums,enumsi}{Example}{Examples}
\crefname{}{example}{examples} % lingmacros \toplabel has no internal name for the kind of label
\Crefname{}{Example}{Examples}
\crefformat{enums}{(#2#1#3)}
\crefformat{enumsi}{(#2#1#3)}
\crefrangeformat{enums}{\mbox{(#3#1#4--#5#2#6)}}
\crefrangeformat{enumsi}{\mbox{(#3#1#4--#5#2#6)}}
\crefformat{}{(#2#1#3)}
\crefname{xnumi}{example}{examples} % gb4e
\crefname{xnumi}{example}{examples} % gb4e
\Crefname{xnumii}{Example}{Examples} % gb4e
\Crefname{xnumii}{Example}{Examples} % gb4e
\crefformat{xnumi}{(#2#1#3)} % gb4e
\crefformat{xnumii}{(#2#1#3)} % gb4e
\crefrangeformat{enums}{\mbox{(#3#1#4--#5#2#6)}}
\crefrangeformat{enumsi}{\mbox{(#3#1#4--#5#2#6)}}
\crefrangeformat{xnumi}{\mbox{(#3#1#4--#5#2#6)}} % gb4e
\crefrangeformat{xnumii}{\mbox{(#3#1#4--#5#2#6)}} % gb4e
\crefmultiformat{enumsi}{(#2#1#3}{, #2#1#3)}{, #2#1#3}{, #2#1#3)}
\crefmultiformat{xnumi}{(#2#1#3}{, #2#1#3)}{, #2#1#3}{, #2#1#3)} % gb4e
\crefmultiformat{xnumii}{(#2#1#3}{, #2#1#3)}{, #2#1#3}{, #2#1#3)} % gb4e
\crefrangemultiformat{enumsi}{(#3#1#4--#5#2#6}{, #3#1#4--#5#2#6)}{, #3#1#4--#5#2#6}{, #3#1#4--#5#2#6)}
\crefrangemultiformat{xnumi}{(#3#1#4--#5#2#6}{, #3#1#4--#5#2#6)}{, #3#1#4--#5#2#6}{, #3#1#4--#5#2#6)} % gb4e
\crefrangemultiformat{xnumii}{(#3#1#4--#5#2#6}{, #3#1#4--#5#2#6)}{, #3#1#4--#5#2#6}{, #3#1#4--#5#2#6)} % gb4e

\ifx\creflastconjunction\undefined%
\newcommand{\creflastconjunction}{, and\nobreakspace} % Oxford comma for lists
\else%
\renewcommand{\creflastconjunction}{, and\nobreakspace} % Oxford comma for lists
\fi%

\newcommand*{\Fullref}[1]{\hyperref[{#1}]{\Cref*{#1}: \nameref*{#1}}}
\newcommand*{\fullref}[1]{\hyperref[{#1}]{\cref*{#1}: \nameref{#1}}}
\makeatletter
\renewcommand{\@BIBLABEL}{\@emptybiblabel}
\makeatother

\newcommand{\ucca}[1]{\textcolor{gray}{\textbf{\textsf{#1}}}}
\newcommand{\sst}[1]{\textsc{#1}}
\newcommand{\lexcat}[1]{\textsl{#1}}

\definecolor{orange}{rgb}{1,0.5,0}
\definecolor{mdgreen}{rgb}{0.05,0.6,0.05}
\definecolor{Acolor}{HTML}{EC5D57} % poppy red
\definecolor{Pcolor}{HTML}{70BF41} % grass green
\definecolor{Scolor}{HTML}{51A7F9} % sky blue
\definecolor{Lcolor}{HTML}{B36AE2} % friendly purple
\definecolor{mdblue}{rgb}{0,0,0.7}
\definecolor{dkblue}{rgb}{0,0,0.5}
\definecolor{dkgray}{rgb}{0.3,0.3,0.3}
\definecolor{slate}{rgb}{0.25,0.25,0.4}
\definecolor{gray}{rgb}{0.5,0.5,0.5}
\definecolor{ltgray}{rgb}{0.7,0.7,0.7}
\definecolor{purple}{rgb}{0.7,0,1.0}
\definecolor{lavender}{rgb}{0.65,0.55,1.0}

\newcommand{\com}[1]{}

\newcommand{\finalversion}[1]{} % hide for now, restore for camera-ready

\usetikzlibrary{shapes,shapes.misc}

\newcommand{\ApplyGradient}[1]{%
  \pgfmathsetmacro{\PercentColor}{(#1-0)/7.58}%
  \pgfmathsetmacro{\PercentInverse}{ifthenelse(\PercentColor > 65, 0, 100)}%
  \edef\x{\noexpand\cellcolor{red!\PercentColor}}\x\textcolor{black!\PercentInverse}{#1}%
}
\newcolumntype{R}{>{\small\collectcell\ApplyGradient}{c}<{\endcollectcell}}

\def\arraystretch{1.1}

\newcommand{\cmark}{\large\color{green}\ding{51}}
\newcommand{\xmark}{\large\color{red}\ding{55}}
\newcommand{\xcmark}{\large\color{orange}\checkmark\kern-1.1ex\raisebox{.7ex}{\rotatebox[origin=c]{125}{--}}}

\newcommand{\emldisplay}[2]{\texttt{\href{mailto:#1}{#2}}}
\newcommand{\eml}[1]{\emldisplay{#1}{#1}}

\colingfinalcopy % Uncomment this line for the final submission

\title{Comparison by Conversion: \\ 
Reverse-Engineering UCCA 
from Syntax and Lexical Semantics
}
\author{
Daniel Hershcovich$^\diamondsuit$ \quad
Nathan Schneider$^\spadesuit$ \quad
Dotan Dvir$^\heartsuit$ \\\bfseries
Jakob Prange$^\spadesuit$ \quad
Miryam de Lhoneux$^\diamondsuit$ \quad
Omri Abend$^\heartsuit$ \\
$^{\diamondsuit}$University of Copenhagen \quad
$^{\spadesuit}$Georgetown University \quad
$^{\heartsuit}$Hebrew University of Jerusalem \\
\ttfamily\{\emldisplay{dh@di.ku.dk}{dh},\emldisplay{ml@di.ku.dk}{ml}\}@di.ku.dk,\quad
\{\emldisplay{nathan.schneider@georgetown.edu}{nathan.schneider},\emldisplay{jp1724@georgetown.edu}{jp1724}\}@georgetown.edu, \\
\ttfamily\eml{dotan.dvir@mail.huji.ac.il},\quad\eml{oabend@cs.huji.ac.il}
}

\date{}

\begin{document}

\maketitle

\begin{abstract}

Building robust natural language understanding systems will require a clear characterization of whether and how various linguistic meaning representations complement each other. To perform a systematic comparative analysis, we evaluate the mapping between meaning representations from different frameworks using two complementary methods: (i)~a rule-based converter, and (ii)~a supervised \emph{delexicalized} parser that parses to one framework using only information from the other as features. We apply these methods to convert the STREUSLE corpus (with syntactic and lexical semantic annotations) to UCCA (a graph-structured full-sentence meaning representation).  Both methods yield surprisingly accurate target representations, close to fully supervised UCCA parser quality---indicating that UCCA annotations are partially redundant with STREUSLE annotations.  Despite this substantial convergence between frameworks, we find several important areas of divergence.
\end{abstract}

\section{Comparing Meaning Representations}\label{sec:introduction}

\blfootnote{
    \hspace{-0.65cm}  % space normally used by the marker
    This work is licensed under a Creative Commons 
    Attribution 4.0 International License.
    License details:
    \url{http://creativecommons.org/licenses/by/4.0/}.
}

% \daniel{Error analysis as in \cite{He2017DeepSR} with rules to recover specific gold annotations}

Several symbolic meaning representations (MRs) support human annotation of text with broad coverage \cite{abend2017state,Oep:Abe:Haj:19}.
To date, it is still not completely clear, for all frameworks, what linguistic semantic phenomena they encode, and how it compares to the content represented by the others.
It therefore behooves us to develop a firm linguistic understanding of MRs.
In particular:
are they merely a coarsening and rearranging of syntactic information, such as is encoded in Universal Dependencies \citep[UD;][]{nivre2016universal,nivre-20}?
To what extent do they take lexical semantic properties into account?
What does this suggest about the potential for exploiting simpler or better-resourced linguistic representations for improved MR parsing?
Intuitively, we ask whether:
\begin{align*}
     \text{sentence-level MR} \stackrel{?}{=} %\\
     \text{syntax} + \text{lexical semantics}
\end{align*}

To address this question, we examine UCCA, a document-level MR often used for sentence-level semantics (see~\cref{sec:ucca}).
\newcite{hershcovich2019content} began to examine the relation of UCCA to syntax, contributing a corpus with gold standard UD and UCCA parses, heuristically aligning them, and quantifying the correlations between syntactic and semantic labels.
% did not go beyond analysis, stopping short of using linguistic mappings in a targeted way for UCCA parsing.
Conversely, \newcite{hershcovich2018multitask} 
provided some initial evidence that other MRs can be brought to bear on the UCCA parsing task via multitask learning, 
but left the details of the relationship between representations to latent (and opaque) parameters of neural models.

In this paper, we aim to close the gap between the two previous investigations by (1) building an interpretable rule-based system to convert from shallower representations
(syntax and lexical semantic units/tags) 
into UCCA, forcing us to be linguistically precise about 
what UCCA captures and how it ``decomposes'';
and (2) training top-performing supervised parsers in a delexicalized setting with
only syntactic and lexical semantic features, as a data-driven mapping
corroborating the rule-based approach.

We perform our analysis on the Reviews section of the English Web Treebank \cite{bies2012english},
which has been manually annotated with UD and UCCA; and STREUSLE for lexical semantics (\S\ref{sec:representations}).
Although at present we only have the necessary evaluation data for English, 
the linguistic representations we examine have been applied to multiple languages (\cref{sec:representations}).
% \daniel{``hedge this a little bit. Even though all these representations may be lexicon-free, they may have language-specific tags (U.D.), or non-trivial alignment between labels across languages.''}
Our approach can thus be applied cross-linguistically with minimal adaptation.
Our contributions are:
\begin{itemize}%[leftmargin=*]
  \item Delexicalized rule-based and supervised UCCA parsers, based only on syntax and lexical semantics.
  \item A linguistically motivated analysis of similarities and differences between the frameworks.
\end{itemize}

% The rest of the paper is organized as follows:
% \Cref{sec:representations} presents UCCA in detail, as well as the
% syntactic (UD) and lexical semantic (STREUSLE) resources used in this work.
% \Cref{sec:converter} presents our rule-based converter,
% and \Cref{sec:parsing} our method for training supervised UCCA parsers
% with lexical semantic features.
% In \cref{sec:experiments} we present our experimental setup,
% and in \cref{sec:results} our results.
% Finally, \cref{sec:discussion} includes a discussion of the various divergences revealed by the experiments, and \cref{sec:conclusion} concludes the paper.\finalversion{\nss{rewrite to focus on content rather than structure}}

%%%%%%%%%%%%%%%%%%%%%%%%%%%%%%%%%%%%%%%%%%%%%%%%%%%%%%%%%%%%%%%%%%%%%%%%%%%%%%%%%

%Omri
\section{Representations under Consideration}\label{sec:representations}

\begin{figure*}[t]\centering
\includegraphics[width=\textwidth]{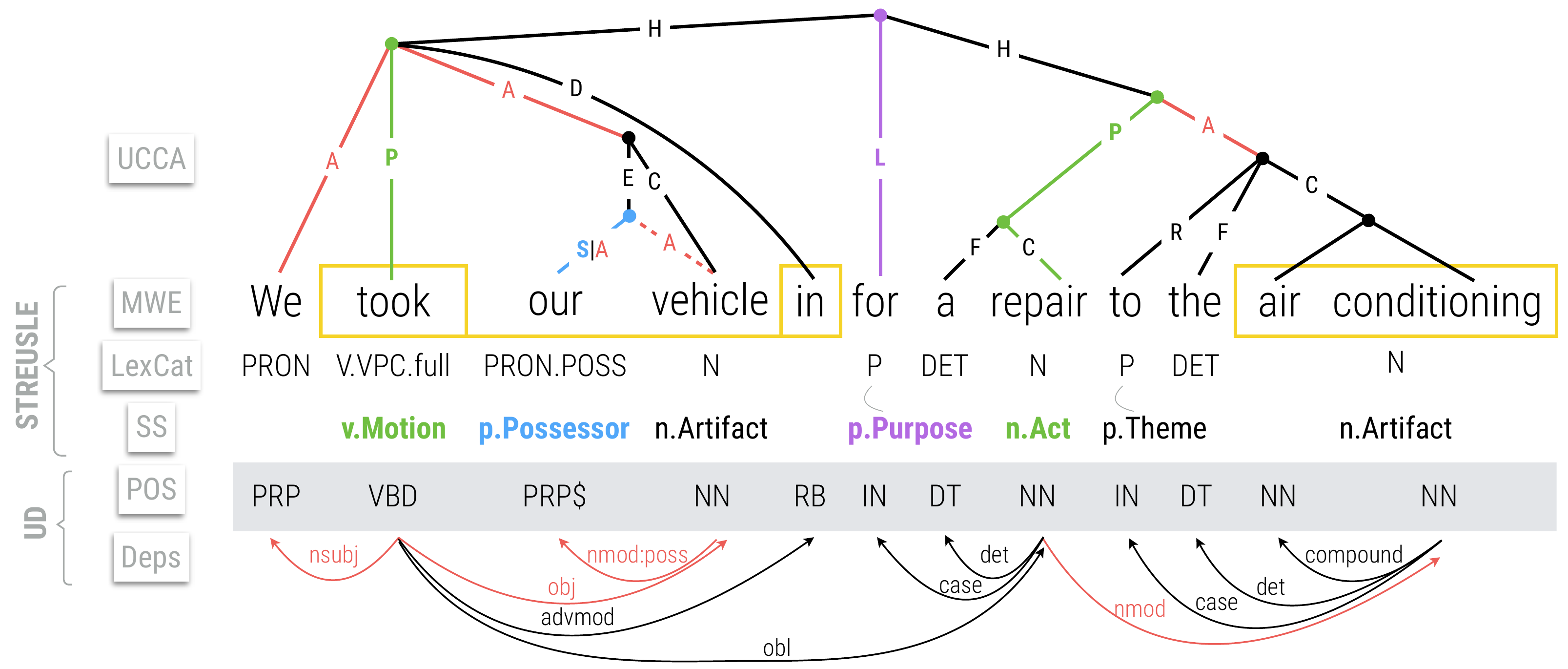}
\caption{Example sentence from the Reviews training set
(\texttt{reviews-086839-0003}, ``We took our vehicle in for a repair to the air conditioning''),
with UCCA, STREUSLE, and UD annotations. \\
$\diamond$ UCCA abbreviations: \ucca{\color{black}H} = parallel scene, \ucca{\color{Lcolor}L} = scene linker, \ucca{\color{Pcolor}P} = process (dynamic event), \ucca{\color{Scolor}S} = state, \ucca{\color{Acolor}A} = scene participant, \ucca{\color{black}D} = scene adverbial, \ucca{\color{black}E} = non-scene elaborator, \ucca{\color{black}C} = center (non-scene head), \ucca{\color{black}R} = relator, \ucca{\color{black}F} = functional element. The STREUSLE and UD part is adapted from \protect\newcite{liu2020lexical}.}
\label{fig:ex}
\end{figure*}

The increasing interest in semantic representation and parsing, and the partial overlap in content between the different frameworks \cite{Oep:Abe:Haj:19}, is a main motivation for our inquiry into content differences between UD and STREUSLE, and UCCA. We expect our inquiry to be relevant to other schemes, both in developing a general methodology, and in the insights gathered. For example, besides STREUSLE, UD also serves as the backbone of the DeComp scheme \cite{white2016universal}, and so information as to its semantic content is important there as well. Argument structural phenomena are at the heart of many MRs, which provide further motivation for empirical studies to the extent lexical semantics and syntax can encode them.

\subsection{Universal Conceptual Cognitive Annotation}\label{sec:ucca}
Universal Conceptual Cognitive Annotation \cite{abend2013universal} targets a level of semantic granularity that abstracts away from syntactic paraphrases in a typologically-motivated, cross-linguistic fashion \cite{sulem2015conceptual}, building on Basic Linguistic Theory \cite{Dixon:basic}, an influential framework for linguistic description. 
The scheme does not rely on language-specific resources, and sets a low threshold for annotator training. Beyond syntactic paraphrases, UCCA encodes lexical semantic properties such as the aspectual distinction between states and processes (whether an event evolves in time or not).

UCCA has been applied to text simplification \cite{sulem2018simple}
and evaluation of text-to-text generation \cite{birch2016hume,choshen2018usim,sulem2018samsa}.
UCCA corpora are available for English, French and German, 
and pilot studies have been conducted on additional languages.
% UCCA parsing has been targeted in two shared tasks \cite{hershcovich2019shared,Oep:Abe:Haj:19}.
Here we summarize the principles and main distinctions in UCCA.\footnote{For further details, see the extensive UCCA annotation manual:\\ \url{https://github.com/UniversalConceptualCognitiveAnnotation/docs/blob/master/guidelines.pdf}}

In UCCA, an analysis of a text passage is a DAG (directed acyclic graph)
over semantic elements called \textbf{units}.
A unit corresponds to (is \textit{anchored} by) one or more tokens,
labeled with one or more semantic \textbf{categories} in relation to a parent unit.\footnote{UCCA also supports \textbf{implicit units} which do not correspond to any tokens \cite{cui2020refining}, 
but these are excluded from parsing evaluation and we ignore them for purposes of this paper.}
The principal kind of unit is a \textbf{scene} denoting a situation mentioned in the sentence, typically involving a scene-evoking \textbf{predicate}, participants, and (perhaps) modifiers. 
Each predicate is labeled as either State (\ucca{\color{Scolor}S}) or Process (\ucca{\color{Pcolor}P}). 
\Cref{fig:ex} contains three scenes: one anchored by the Process \textit{took}; one anchored by the Process \textit{a repair}; and one anchored by the possessive pronoun \textit{our}, which indicates a stative possession relation.
A Participant (\ucca{\color{Acolor}A}) of a scene is typically an entity or location involved. 
Adverbials (\ucca{D}) modify scenes with respect to properties like negation, modality, causativity, direction, manner, etc., which do not constitute an independent situation or entity.
Temporal modifiers are labeled Time (\ucca{T}).

Scenes in UCCA can relate to one another in one of three ways.
A Scene can serve as a Participant within a larger scene; 
a Scene can serve to elaborate on a Participant within a Scene (typically relative clauses);
or scenes can be related by \textbf{parallel linkage} in a unit that consists of Parallel Scenes (\ucca{H}) and possibly Linkers (\ucca{\color{Lcolor}L}) describing how they are related. This is seen at the top level of \cref{fig:ex}, where the taking and repair scenes are parallel and the purposive \textit{for} is a linker.

Other categories only apply to units with no predicate: a semantic head---the Center (\ucca{C}); modifiers of Quantity (\ucca{Q}); and other modifiers, called Elaborators (\ucca{E}). 
An Elaborator may itself be a scene, as in \textit{our vehicle}, where the scene of possession elaborates on the vehicle entity. Similarly, \textit{blue vehicles} would be analyzed with a stative scene of blueness that elaborates on the vehicles. %in question.

Apart from the main semantic content of scenes and participants, UCCA provides the categories: Relator (\ucca{R}) for grammatical markers expressing how a unit relates to its parent unit---in English, these are mainly prepositions and the possessive \textit{'s}; Function (\ucca{F}) for other grammatical markers with minimal semantic content, such as tense auxiliaries, light verbs, and articles.
Other categories are used for expressing coordination and for expressions expressing speaker perspective outside the propositional structure of the sentence.
%The least contentful elements---\ucca{F}s, and to a lesser extent \ucca{R}s---are subject to considerable variation when paraphrasing or translating a sentence. 
%For instance, consider that \textit{for$_{\ucca{\color{Lcolor}L}}$ [a repair to the air conditioning]$_{\ucca{H}}$} can be paraphrased as \textit{[in order to]$_{\ucca{\color{Lcolor}L}}$ [get the air conditioning repaired]$_{\ucca{H}}$}, which omits an article (\ucca{F}) and a preposition (\ucca{R}).
%Punctuation tokens are attached to units as \ucca{U} in the data, but are excluded from standard UCCA parsing evaluation.
Semantically opaque multi-word expressions (e.g., \textit{air conditioning} in \cref{fig:ex}) are called \textbf{unanalyzable units},
and are not analyzed internally.
UCCA distinguishes {\bf primary edges} that always form a tree, and \textbf{remote edges}, which express reentrancies, such as the dotted edge from the possession scene unit to \textit{vehicle}.

\subsection{Universal Dependencies}\label{sec:ud}

UD is a syntactic dependency scheme used in many languages,
aiming for cross-linguistically consistent and coarse-grained treebank
annotation. Formally, UD uses bi-lexical trees, with edge labels 
representing syntactic relations.
An example UD tree appears at the bottom of \cref{fig:ex}.

%  One aspect of UD similar to UCCA is its preference of lexical (rather than functional) heads.
%  For example, in auxiliary verb constructions (e.g., ``is eating''), UD
%  marks the lexical verb (\textit{eating}) as the head, while other dependency schemes
%  may select the auxiliary \textit{is} instead.
%  While the approaches are largely inter-translatable
%  \newcite{Schwartz:12}, lexical head schemes are more similar in form to semantic schemes,
%   such as UCCA and semantic dependencies \newcite{oepen2016towards}.
   
%UD relations will be written in \texttt{typewriter} font.

\subsection{STREUSLE}\label{sec:streusle}
STREUSLE (Supersense-Tagged Repository of English with a Unified Semantics for Lexical Expressions)
is a corpus annotated comprehensively for several forms of lexical semantics \cite{schneider-15,schneider-18}.
All kinds of \textbf{multi-word expressions} (MWEs) are annotated, giving each sentence a lexical semantic segmentation.\footnote{STREUSLE distinguishes strong MWEs, which are opaque (noncompositional) or idiosyncratic in meaning, and weak MWEs, which represent looser collocations that are nevertheless semantically compositional, like ``highly recommended''.}
Syntactic and semantic tags are then applied
to individual units (single- and multi-word). 
The semantic tags are \textbf{supersenses} for noun, verb, and prepositional\slash possessive units.
Preposition supersenses include two tiers of annotation: \textbf{scene role} labels represent the semantic role of the prepositional \textit{phrase} marked by the preposition, and \textbf{function} labels represent the lexical contribution of the \textit{preposition} in itself. The two labels are drawn from the same supersense inventory and are identical for many tokens.

The \textbf{lexcat} annotations (syntactic category of lexical unit) is a slight extension to the Universal POS tagset, adding categories for certain MWE subtypes, such as light verb constructions, following \newcite{walsh-18} and idiomatic PPs; it also distinguishes possessive pronouns, the possessive clitic \textit{'s}, and discourse expressions.\footnote{STREUSLE tagset documentation:
\url{https://github.com/nert-nlp/streusle/blob/master/CONLLULEX.md}}
\Cref{fig:ex} illustrates the MWE, lexcat, and supersense layers.

STREUSLE itself is limited to English, but many of its component annotations have been applied to other languages: verbal multi-word expressions \cite{parseme1.1}, noun and verb supersenses \cite{picca-08,qiu-11,schneider-13,martinez_alonso-15,hellwig-17}, and preposition supersenses \cite{hwang-17,peng-20,hwang-20}.
% \daniel{COLING guidelines: new dataset(s) used in a paper must be well described, with complete information about how the data were gathered and annotated. Also, assess whether the dataset(s) is made easy to access and has cleared distribution rights}

\newcite{liu2020lexical} presented a comprehensive lexical semantic tagger for STREUSLE, which predicts the comprehensive lexical semantic analysis from text, and is freely available.
\newcite{prange2019made} proposed several procedures for integrating STREUSLE supersenses directly into UCCA, refining its coarse-grained categories with \textit{preposition} supersenses. Enriching a supervised UCCA parser with preposition supersense features from STREUSLE---and, even more so, training a parser to predict supersenses jointly with UCCA---improved parsing performance, revealing the two frameworks to be overlapping but complementary.
% \jp{I think this is good, just it may be confusing that we mention ``STREUSLE supersenses'' in general at first while we only dealt with preposition supersenses. maybe italicize first mention of \textit{preposition}?}

%%%%%%%%%%%%%%%%%%%%%%%%%%%%%%%%%%%%%%%%%%%%%%%%%%%%%%%%%%%%%%%%
\subsection{Related Representations}\label{sec:related_work}

The above annotation schemes define finite inventories of coarse-grained categories to avoid depending on language-specific lexical resources, and thus can in principle be applied to any language. 
This fact distinguishes UCCA and STREUSLE from finer-grained sentence-structural representations like FrameNet \cite{Baker:98,fillmore-baker-09} and the Abstract Meaning Representation \cite{banarescu2013abstract}, which relies on PropBank \cite{Palmer:05}. The Prague Dependency Treebank tectogrammatical layer \cite{bohmova2003prague} uses few lexicon-free roles, but its semantics is determined by a valency lexicon. 
% \daniel{``reference to other work on semantic annotation'',
% ``awareness of the importance of interoperability in (semantic) annotation''}

% More closely related to our approach is the 
The Parallel Meaning Bank \cite{abzianidze2017parallel} uses lexicon-free\footnote{While the lexical items comprising a linguistic utterance
are naturally essential to its meaning, and therefore influence its semantic representation,
by \textit{lexicon-free} we mean the \textit{ontology} and label set of the representation
are not tied to a lexicon or a particular language.} VerbNet \cite{kipper2005verbnet} semantic roles. The STREUSLE tagset for preposition supersenses generalizes VerbNet's role set
to cover non-core arguments/adjuncts of verbs, as well as prepositional complements of nouns and adjectives.
Universal Decompositional Semantics (DeComp) defines semantic roles as bundles of lexicon-free features.
%, inspired by Dowty's theory of proto-roles. 
Cross-linguistic applicability in this case is delegated to the parser, which parses sentences in other languages to their corresponding {\it English} semantics \cite{zhang2018cross}.

%\nss{Something about SDP?}\oa{SDP is really three different schemes, but I think tectogrammatical is similar to UCCA in this respect; the difference is that UCCA has a design principle of stability to translations, and Sulem et al. (2015) shows supporting evidence.}\nss{maybe mention decomp here too}\oa{decomp is indeed lexicon-free, but annotation was only done in English afaik; that said, they do have cross-lingual parsers, so their idea of language generality is automatically mapping text in other languages to English: see Cross-lingual Decompositional Semantic Parsing / EMNLP 2018}

%%%%%%%%%%%%%%%%%%%%%%%%%%%%%%%%%%%%%%%%%%%%%%%%%%%%%%%%%%%%%%%%%%%%%%%%%%%%%%%%%%%%%%%%%%%%%%

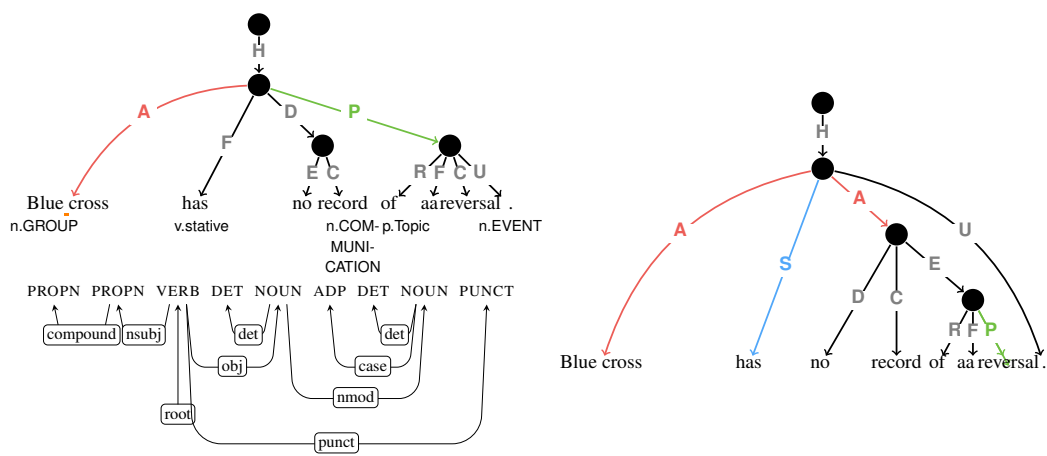
\begin{figure}[ht]
\centering
\begin{minipage}{.45\textwidth}
\centering\setulcolor{orange}\setul{0.5ex}{0.3ex}
\scalebox{.67}{
\begin{tikzpicture}[->,level distance=12mm,line width=1pt,
  level 2/.style={sibling distance=25mm},
  level 3/.style={sibling distance=8mm},
  every circle node/.append style={fill=black},
  every node/.append style={text height=.6ex,text depth=0}]
  \tikzstyle{word} = [font=\rmfamily,color=black]
  \node (1_1) [circle] {}
  {
  child {node (1_2) [circle] {}
    {
    child {node {} {
      child {node (1_4) [word] {\ul{Blue cross}}  edge from parent [draw=none]}
    } edge from parent [draw=none]}
    child {node {} {
      child {node (1_5) [word] {has}  edge from parent [draw=none]}
    } edge from parent [draw=none]}
    child {node (1_6) [circle] {}
      {
      child {node (1_11) [word] {no}  edge from parent node[midway, fill=white]  {\ucca{E}}}
      child {node (1_12) [word] {record}  edge from parent node[midway, fill=white]  {\ucca{C}}}
      } edge from parent node[midway, fill=white]  {\ucca{D}}}
    child [draw=Pcolor] {node (1_7) [circle] {}
      {
      child [draw=black] {node (1_8) [word] {of}  edge from parent node[midway, fill=white]  {\ucca{R}}}
      child [draw=black] {node (1_9) [word] {aa}  edge from parent node[midway, fill=white]  {\ucca{F}}}
      child [draw=black] {node (1_10) [word] {reversal}  edge from parent node[midway, fill=white]  {\ucca{C}}}
      child [draw=black] {node (1_13) [word] {.}  edge from parent node[midway, fill=white]  {\ucca{U}}}
      } edge from parent node[midway, fill=white]  {\ucca{\color{Pcolor}P}}}
    } edge from parent node[midway, fill=white]  {\ucca{H}}}
  };
  \draw[bend right,draw=Acolor] (1_2) to node[midway, fill=white] {\ucca{\color{Acolor}A}} (1_4);
  \draw (1_2) to node[midway, fill=white] {\ucca{F}} (1_5);
\end{tikzpicture}}

{\sf\tiny
\begin{tabular}{cccccccccccc}
n.GROUP &&& v.stative &&& n.COM- & \hspace{-4mm}p.Topic && \hspace{-2mm}n.EVENT \\
&&&&&& MUNI- \\
&&&&&& CATION
\end{tabular}}
\vspace{-3mm}

\scalebox{.8}{
\begin{dependency}[edge below]
% sent_id = reviews-003418-0006
% text = Blue cross has no record of aa reversal.
\begin{deptext}[opacity=0]
Blue  \& cross \& has  \& no  \& record \& of  \& aa  \& reversal \& .     \\
\end{deptext}
\begin{deptext}[font=\small\sc]
propn \& propn \& verb \& det \& noun   \& adp \& det \& noun     \& punct \\
\end{deptext}
\depedge{2}{1}{compound}
\depedge{3}{2}{nsubj}
\deproot{3}{root}
\depedge{5}{4}{det}
\depedge{3}{5}{obj}
\depedge{8}{6}{case}
\depedge{8}{7}{det}
\depedge{5}{8}{nmod}
\depedge[edge unit distance=1em]{3}{9}{punct}
\end{dependency}}
\end{minipage}
\begin{minipage}{.45\textwidth}
\scalebox{.67}{
\begin{tikzpicture}[->,level distance=13mm,line width=1pt,
  level 2/.style={sibling distance=29mm},
  level 3/.style={sibling distance=15mm},
  level 4/.style={sibling distance=7mm},
  every circle node/.append style={fill=black},
  every node/.append style={text height=.6ex,text depth=0}]
  \tikzstyle{word} = [font=\rmfamily,color=black]
  \node (1_1) [circle] {}
  {
  child {node (1_2) [circle] {}
    {
    child {node {} {
    child {node {} {
      child {node (1_4) [word] {Blue cross}  edge from parent [draw=none]}
    } edge from parent [draw=none]}
    } edge from parent [draw=none]}
    child {node {} {
    child {node {} {
      child {node (1_3) [word] {has}  edge from parent [draw=none]}
    } edge from parent [draw=none]}
    } edge from parent [draw=none]}
    child [draw=Acolor] {node (1_5) [circle] {}
      {
      child {node {} {
        child {node (1_8) [word] {no}  edge from parent [draw=none]}
      } edge from parent [draw=none]}
      child {node {} {
        child {node (1_6) [word] {record}  edge from parent [draw=none]}
      } edge from parent [draw=none]}
      child [draw=black] {node (1_9) [circle] {}
        {
        child {node (1_11) [word] {of}  edge from parent node[midway, fill=white]  {\ucca{R}}}
        child {node (1_12) [word] {aa\;{\color{white}a}}  edge from parent node[midway, fill=white]  {\ucca{F}}}
        child {[draw=Pcolor] node (1_10) [word] {reversal}}
        } edge from parent node[midway, fill=white]  {\ucca{E}}}
      } edge from parent node[midway, fill=white]  {\ucca{\color{Acolor}A}}}
    child {node {} {
    child {node {} {
      child {node (1_7) [word] {.}  edge from parent [draw=none]}
    } edge from parent [draw=none]}
    } edge from parent [draw=none]}
    } edge from parent node[midway, fill=white]  {\ucca{H}}}
  };
  \draw[bend right,draw=Acolor] (1_2) to node[midway, fill=white] {\ucca{\color{Acolor}A}} (1_4);
  \draw[draw=Scolor] (1_2) to node[midway, fill=white] {\ucca{\color{Scolor}S}} (1_3);
  \draw[draw=Pcolor] (1_9) to node[midway, fill=white] {\ucca{\color{Pcolor}P}} (1_10);
  \draw[bend left] (1_2) to node[midway, fill=white] {\ucca{U}} (1_7);
  \draw (1_5) to node[midway, fill=white] {\ucca{D}} (1_8);
  \draw (1_5) to node[midway, fill=white] {\ucca{C}} (1_6);
\end{tikzpicture}}
\end{minipage}
\caption{Example sentence (\texttt{reviews-003418-0006}, reading ``Blue cross has no record of aa[\textit{sic}] reversal'') with gold-standard UCCA graph;
STREUSLE MWEs %, lexcat 
and supersenses; and UD coarse-grained POS tags and relations (left);
and UCCA graph output by the rule-based converter (right).\label{fig:ex_rule_ss}}
\end{figure}

\section{First Conversion Approach: Rule-based UCCA Parsing from Syntax and Lexical Semantics}\label{sec:converter}

Here we describe a system to produce UCCA analyses for text,
given UD syntactic graphs and STREUSLE lexical semantic annotation.
%
% The UD-to-UCCA converter of \newcite{hershcovich2019content}\footnote{\nss{URL}}
% was used for assimilating UD and UCCA graphs, to allow comparing their content.
% It attached nodes according to preorder traversal of the original dependency graph,
% keeping the UD dependency relations.
% The remaining differences are cases that could not be disambiguated based on UD alone.
% We extend the converter to account for these differences,
% integrating lexical semantics into the process, and additionally assigning UCCA categories
% to edges.
% We normalize deterministic differences between the structures to abstract away
% from formal differences, as we are interested in a comparison in terms of
% content and semantic distinctions.
%
%
This is a rule-based converter that inspects the UD structure in tandem with STREUSLE annotations to build an UCCA parse.\footnote{The new conversion and analysis code is available at \url{https://github.com/danielhers/streusle/tree/streusle2ucca}}
An analysis of the converter's successes and failures (\cref{sec:discussion})
will, in turn, reveal the similarities and differences between the schemes. What follows is an overview of the algorithm; futher details about the rules are given in \cref{sec:rules}.\footnote{Before writing the new converter from the ground up, we tried modifying the \newcite{hershcovich2019content} UD-only conversion code to use STREUSLE information. Details and analysis of that system appear in \cref{sec:altconverter}.}

\paragraph{MWEs and unanalyzable units.}
Based on STREUSLE MWE annotation,
we group together text tokens into single unanalyzable semantic units when
they are annotated as strong MWEs (we do not use weak MWEs, as UCCA does not encode them),
except for light verbs and annotations that would lead to cycles.
MWE-internal dependency edges are discarded so they will not be
processed later.
Punctuation is marked \texttt{U}.
Mappings between units and dependency nodes are maintained.
For example, in
\begin{quote}
This is one of the worst places I have stayed, we {\bf cut} out[\textit{sic}] stay {\bf short} and went to the Mulberry. (\texttt{reviews-023620-0001}),
\end{quote}
\textit{cut\dots short} is an unanalyzable unit according to the gold UCCA annotation,
but the syntactic relation between
\textit{cut} and \textit{short}, namely \texttt{xcomp},
does not indicate that; only few
\texttt{xcomp}s, in general, correspond to UCCA unanalyzable units
\cite{hershcovich2019content}.
In STREUSLE, however, \textit{cut\dots short} is annotated as a strong MWE, 
whose lexcat is \lexcat{V.VID} (idiomatic verb).
Our rules create an unanalyzable unit covering this phrase,
which matches the gold UCCA annotation in this case.
The same is true for \textit{Blue cross}\footnote{Short for \textit{Blue Cross Blue Shield}, a well-known health insurance organization in the United States.} in \cref{fig:ex_rule_ss}.

\paragraph{STREUSLE supersenses and UCCA scene-evoking phrases.}
In a top-down traversal of the dependency parse, we visit each word's
lexical unit and decide whether it evokes a main relation (scene-evoking phrase), using rules based on
syntactic and lexical semantic features:
copular \emph{be} and stative \emph{have},
adjectives (excluding a small list of quantity adjectives),
existential \emph{there},
non-discourse adverbs with a copula dependent,
predicative prepositions and
copulas introducing a predicate nominal,
as well as common nouns supersense-tagged as \sst{n.attribute}, \sst{n.feeling}, or \sst{n.state}
are labeled \ucca{\color{Scolor}S};
verbs,
\emph{thanks}, \emph{thank you} and common nouns supersense-tagged as \sst{n.act}, \sst{n.phenomenon}, \sst{n.process}, or \sst{n.event} (with the exception
of nouns denoting a part of the day) are labeled \ucca{\color{Pcolor}P}.
For example, the noun \textit{reversal} in \cref{fig:ex_rule_ss}
is marked as \ucca{\color{Pcolor}P},
since its supersense of \sst{n.event} informs us that this is
a scene-evoking noun.
Relational nouns, supersense-tagged as \sst{n.person} or \sst{n.group} that also
match kinship/occupation lists or suffixes, are labeled \ucca{\color{Pcolor}P}+\ucca{\color{Acolor}A}.

\paragraph{STREUSLE lexical categories and UCCA edge categories.}
After identifying main relations and non-scene units, the exact category for some units still needs to be determined. The main relations themselves are labeled
\ucca{\color{Pcolor}P} or \ucca{\color{Scolor}S}
depending on lexical categories and supersenses.
Determiners, auxiliaries, and copulas are generally labeled \ucca{F}; vocatives
and interjections, \ucca{G}. Exceptions include modal auxiliaries
(\ucca{D}), and demonstrative/quantifier determiners modifying a non-scene unit
(\ucca{E}/\ucca{Q} respectively).
Verbal arguments and modifiers of scene-evoking non-verbal phrases are labeled
\ucca{\color{Acolor}A}.
Possessive clitics and prepositions are attached as \ucca{R}, unless
a possessive clitic marks canonical possession, in which case it is
\ucca{\color{Scolor}S}.

\paragraph{Secondary verb constructions.}
These constructions are structured differently in UCCA and UD:
in ``members who won't stop talking'',
the verb ``stop'' is the UD head and ``talking'' is a dependent.
In UCCA ``stop'' is \ucca{D} and ``talking'' is the main
relation, labeled \ucca{\color{Pcolor}P}.
To normalize the treatment of these constructions,
they are marked and eventually restructured such that the syntactic head is labeled
\ucca{D} and the syntactic dependent is labeled as the main relation.

\paragraph{Coordination and lexical heads.}
Traversing the graph top-down again, coordination between
scene units is labeled \ucca{L} (Linker), and between
non-scene units, \ucca{N} (Connector).
Scene units are labeled \ucca{H},
and non-scene unit heads, \ucca{C}.
Lexical
heads of units are labeled \ucca{C}, \ucca{\color{Pcolor}P}, or \ucca{\color{Scolor}S}, and
scene units as \ucca{H} where necessary;
in ``X of Y'' constructions involving
quantities/species, Y is identified as the \ucca{C}.

%%%%%%%%%%%%%%%%%%%%%%%%%%%%%%%%%%%%%%%%%%%%%%%%%%%%%%%%%%%%%%%%%%%%%%%%%%%%%%%
\section{Second Conversion Approach: Delexicalized Supervised UCCA Parsing}\label{sec:parsing}

%The relation between UCCA's distinctions and lexical and syntactic ones,
%revealed in \cref{sec:converter}, suggest several methods for improving UCCA parsing.

Previous work tackled the UCCA parsing task using supervised learning.
In order to complement and validate the analysis of the rule-based converter, we  compare its findings to a delexicalized supervised parser, that can be seen as inducing a converter from data.
%we consider leading UCCA parsers as near-optimal mappings from lexical input to UCCA graphs.
By removing all word and lemma features from these parsers, and instead adding features based on gold UD and STREUSLE annotations, we obtain supervised ``converters'', which can be used for data-driven analysis and complement the rules.
% We also test a combination of all features.

%\paragraph{Rule-based parsing.}
%The parser itself serves as a rule-based UCCA parser,
%which requires UD and STREUSLE annotation as input.

\paragraph{TUPA.}

This UCCA parser \cite{hershcovich2017a}
is based on a transition-based algorithm with a neural network
transition classifier, using a BiLSTM for encoding input representation,
with word, lemma, and syntactic features embedded as real-valued features.
% Adding lexical semantic features to TUPA
% can facilitate its ability to learn UCCA concepts.
We add the supersense and lexcat from STREUSLE as
embedding inputs to the TUPA BiLSTM
(concatenated with existing inputs).
For prepositions, we add both the scene role and function 
(see~\cref{sec:representations}).\footnote{Our code to enrich UCCA data with STREUSLE features is available at \url{https://github.com/jakpra/ucca-streusle}}
% \jp{this is all accurate (TODO: rewrite a little bit to make more concise and easier to read)}

% \paragraph{Features based on the rule-based parser.}
% Lexical semantics can be used even more readily
% by combining TUPA with the rule-based parser (\cref{sec:converter}).
% We add the following features for each token, based on the
% rule-based parser's output:
% depth (distance from the root), and
% depth and category of lowest common ancestor with the next token.

\paragraph{HIT-SCIR Parser.}
This is a transition-based parser for several MR frameworks, including UCCA \cite{che-etal-2019-hit}. 
It achieved the highest average score in the CoNLL 2019 shared task \cite{Oep:Abe:Haj:19}.
% using stack LSTMs \cite{dyer-etal-2015-transition} to represent the stack, the buffer, and the sequence of actions. These representations are updated after every transition.
While \newcite{che-etal-2019-hit} fine-tuned BERT \cite{devlin-etal-2019-bert} for contextualized word representation, 
our delexicalized version replaces it with UD and STREUSLE features: POS tag, dependency relation, supersenses (scene role and function; see~\cref{sec:representations}), the lexical category of the word or the MWE that the word is part of, and the BIO tag. These are concatenated to form word representations.\footnote{Our modified code for the HIT-SCIR parser is available at \url{https://github.com/danielhers/hit-scir-ucca-parser}}

%%%%%%%%%%%%%%%%%%%%%%%%%%%%%%%%%%%%%%%%%%%%%%%%%%%%%%%%%%%%%%%%%%%%%%%%%%%%%%%

% \begin{wraptable}{r}{6.5cm}
% \vspace{-2cm}
\begin{table}[t]
\centering
\begin{tabular}{lccc}
& \bf Training & \bf Development \\
\hline
\# Tokens & 44,804 & 5,394\\
\# Sentences & \hphantom{0}2,723 & \hphantom{0,}554 \\
\end{tabular}
\caption{EWT Reviews data statistics. We use only the training and development splits for analysis.\label{tab:data}}
\end{table}
% \end{wraptable}

\section{Experiments}\label{sec:experiments}

% \begin{enumerate}
%     \item Baselines
%     \begin{enumerate}
%         \item TUPA \cite{hershcovich2017a}
%         \item TUPA+MT \cite{hershcovich2018multitask}
%         \item TUPA+BERT \cite{Her:Arv:19}
%         \item HLT-SUDA \cite{jiang-etal-2019-hlt}
%         \item HIT-SCIR \cite{che-etal-2019-hit}
%     \end{enumerate}
%     \item For each baseline,
%     \begin{enumerate}
%         \item Oracle transformations (gold UCCA) for development \cite{He2017DeepSR}
%         \begin{enumerate}
%             \item Fix label where yield is correct (equivalent to unlabeled F1)
%             \item Merge spans
%             \item Split spans
%             \item Fix boundary
%             \item Drop primary child
%             \item Add primary child
%             \item Drop remote child
%             \item Add remote child
%         \end{enumerate}
%         \item Oracle transformations (gold UD) for development
%         \item Oracle transformations (gold STREUSLE) for development
%         \item Predicted transformations (UD+STREUSLE)
%         \begin{enumerate}
%             \item for development
%             \item for test
%         \end{enumerate}
%     \end{enumerate}
% \end{enumerate}

\paragraph{Data.}

We use the Reviews section from UD~2.6 English\_EWT \cite{11234/1-3226},
with lexical semantic annotations from STREUSLE~4.4
\cite{schneider-15,schneider-18},\footnote{\url{https://github.com/nert-nlp/streusle}}
% \daniel{Use STREUSLE tagger}\footnote{\url{https://github.com/nelson-liu/streusle-tagger}}
and with UCCA graphs from UCCA\_English-EWT~v1.0.1
\cite{hershcovich2019content}.\footnote{\url{https://github.com/UniversalConceptualCognitiveAnnotation/UCCA_English-EWT}}
We use the standard train/development split for this dataset,
and do not use the test split to avoid over-analyzing it,
although all datasets contain annotations for it too.
% \footnote{We
% do not experiment with other UCCA data despite it being available, as
% it is based on an older version of the UCCA guidelines; as the rules and the
% Reviews annotated corpus are both using UCCA v2, it would be an unfair comparison.}
The data statistics are listed in \cref{tab:data}.

\paragraph{Rule-based converters.}

% We additionally visualize TUPA's learning curve
% by training until convergence only on $\{100,200,\ldots,2700\}$ sentences (\cref{fig:learning_curve}).\daniel{``How were these chosen? Supposedly this choice might have an effect on the performance?''}

% We also evaluate a multitask version of TUPA,
% with UD parsing (on EWT 2.5) as an auxiliary task,
% providing additional training data and assisting in
% learning relevant generalizations.
% This has been shown to improve UCCA parsing,
% especially with little
% training data \cite{hershcovich2018multitask}.

We evaluate the rule-based converter (\cref{sec:converter}),
as well as the syntax-based converter from
\newcite{hershcovich2019content}, which uses the UD tree and a majority-based
category mapping (based on the most common UCCA category in the training set for each UD relation).
This converter is oblivious to lexical semantics.

\paragraph{Parsers.}

We train TUPA v1.3 %\footnote{\url{https://github.com/danielhers/tupa}}
and the HIT-SCIR parser %\footnote{\url{https://github.com/DreamerDeo/HIT-SCIR-CoNLL2019}}
with gold-standard features from UD and STREUSLE, for equal conditions
with the converters,
using default hyperparameters.
Categorical features are added as 20-dimensional embeddings.
% In addition to delexicalized parsing (with UD \& STREUSLE, no words),
% for comparison, we evaluate the parsers with 100-dimensional word2vec \cite{mikolov2013efficient}
% word embeddings
% for TUPA and BERT-Large for HIT-SCIR (no UD \& STREUSLE, with words).
% We also evaluate the parsers with both word feature and the syntactic
% and lexical semantic features (with UD \& STREUSLE, with words),
% assessing whether they complement each other in terms of parsing performance.
Scores are averaged over 3 models with different random seeds.
For TUPA, we ablate UD- or STREUSLE-based features to quantify the contribution of each.

% \paragraph{Ablations.}

% We conduct ablation experiments for the rule-based parser,
% removing specific rules
% to test their effect on the overall performance:
% supersense rules for some or all words;
% MWE information;
% gold UD, using instead parsed UD (with gold tokenization and POS)
% by UDPipe 1.2 \cite{11234/1-2998};
% lemma-list based rules (lexicons);
% and UCCA postprocessing.

\paragraph{Evaluation.}

We use standard UCCA parsing evaluation, matching edges by the terminal yields of their endpoint units.\footnote{The terminal yield of a unit is defined based on the
graph's primary edges only, as standard in UCCA evaluation.}
Labeled precision, recall and F1-score consider the edge categories when matching
edges.
Where an edge has multiple categories,
each of them is considered separately.

% \begin{wraptable}{r}{8cm}
\begin{table}[t]
% \vspace{-1cm}
\centering
\begin{tabular}{@{}lcc@{}}
& \small\bf Primary F1 & \small\bf Remote F1 \\
\midrule
Syntax-based converter with gold UD \cite{hershcovich2019content} & 56.6 & 28.0 \\
Our rule-based converter with gold UD + STREUSLE & 71.7 & 44.2 \\
\midrule
TUPA, delexicalized with gold UD + STREUSLE features       & 69.5 & 46.4 \\
\hphantom{2em} UD features only       & 64.4 & 35.9 \\
\hphantom{2em} STREUSLE features only & 62.4 & 27.5 \\[3pt]
HIT-SCIR, delexicalized with gold UD + STREUSLE features &  67.9 & 41.6\\
\midrule
TUPA with gold UD features + GloVe \cite{hershcovich2019content} & 71.7 & 47.0\\
HIT-SCIR (BERT-Large) &  71.9 & 41.8\\
% \hphantom{2em} with gold UD + STREUSLE features &  70.1 & 37.0\\
HIT-SCIR (GloVe) &  67.0 & 42.4\\
\hphantom{2em} with gold UD + STREUSLE features &  72.2 & 46.9\\
% https://docs.google.com/spreadsheets/d/1qSljskZhWlAQvHDZsbT-zhUrit0wzQtyIxmUQDpYYP4/edit#gid=445422108
% https://docs.google.com/spreadsheets/d/1KSg86VZ9oO21e2IdGlz73yEh-mYcThShTr4URx80Tgk/edit#gid=0
\end{tabular}
\caption{Labeled F1 (in~\%) for primary and
remote edges on the UCCA EWT Reviews dev set,
for rule-based systems (top),
delexicalized supervised parsers with gold UD+STREUSLE (middle),
and supervised parsers with word features (bottom).
% Syntax-based converter and TUPA (gold UD + GloVe) are from \protect\newcite{hershcovich2019content}.
\label{tab:results}}
\end{table}
% \end{wraptable}

\section{Results}\label{sec:results}

\Cref{tab:results} shows the EWT Reviews dev scores.
For comparison with parsers that have access to words,
we also show the TUPA dev results from
\newcite{hershcovich2019content}, who used syntactic features from
the gold UD annotation and GloVe \cite{pennington2014glove};\footnote{Parsing from gold features is by no means a realistic scenario, but we give the scores as a reference for the converters.}
and the HIT-SCIR parser with
BERT/GloVe, and with UD+STREUSLE features.

% , and without any structured signal in the input the parser cannot do much more than guess. \jp{I'm not super happy with this last sentence---it seems too engineering-y and not enough linguisticky... I'll look at it again tomorrow :)}

Rules with gold UD and STREUSLE close the gap
between the syntax-based converter and parsers with word information,
reaching the same primary labeled F1 as TUPA with word features.
This is surprising (since supervised parsers are known to usually outperform rule-based ones),
and suggests that the training data (see \cref{tab:data}) was insufficient for the parser
to learn a mapping as accurate as the complex conversion rules (described in \S\ref{sec:converter}).
Enhancing GloVe-based HIT-SCIR with UD and STREUSLE yields similar results.
However, many errors remain in both approaches, indicating that UCCA and STREUSLE are
\textit{far from equivalent}.
We analyze these errors in \S\ref{sec:discussion} to investigate the frameworks
and the relationship between them.

\paragraph{Ablations.}
Noticeable drops in the ablations (TUPA with UD\slash STREUSLE only) show that both UD-provided structure and relation\slash entity types from STREUSLE supersenses are needed to make up for the missing lexical information, but also that lacking UD hurts more.
This is expected, as the parser resorts to guessing when it lacks sufficiently informative input, and the chance of errors when guessing the UCCA \textit{structure} (for which UD is informative) is much larger than for assigning edge labels (for which STREUSLE provides more fine-grained cues).
The ablated TUPA models still outperform the syntax-based converter, indicating that there are indeed structural signals in STREUSLE and semantic signals in UD, which TUPA can salvage.

\begin{table*}[t]
\centering
\setlength\tabcolsep{4.8pt}
\begin{tabularx}{\textwidth}{@{}lRRRRRRRRRRRRRRRRRRR@{}}
\multicolumn{5}{@{}l}{Predicted Category} & \multicolumn{15}{c}{Gold Category} \\
&\multicolumn{1}{c}{\ucca A}&\multicolumn{1}{c}{\ucca {A|G}}&\multicolumn{1}{c}{\ucca {A|P}}&\multicolumn{1}{c}{\ucca {A|S}}&\multicolumn{1}{c}{\ucca C}&\multicolumn{1}{c}{\ucca D}&\multicolumn{1}{c}{\ucca {D|T}}&\multicolumn{1}{c}{\ucca E}&\multicolumn{1}{c}{\ucca F}&\multicolumn{1}{c}{\ucca G}&\multicolumn{1}{c}{\ucca H}&\multicolumn{1}{c}{\ucca L}&\multicolumn{1}{c}{\ucca N}&\multicolumn{1}{c}{\ucca P}&\multicolumn{1}{c}{\ucca Q}&\multicolumn{1}{c}{\ucca R}&\multicolumn{1}{c}{\ucca S}&\multicolumn{1}{c}{\ucca T}&\multicolumn{1}{c}{\ucca{$\emptyset$}}\\
\ucca {A}&758&4&7&12&17&11&&9&4&1&6&1&&14&1&1&19&&150\\
\ucca {A|P}&&&&1&1&&&&&&&&&&&&&&\\
\ucca {A|S}&&&&8&2&&&&&&&&&&&&&&\\
\ucca {C}&50&&7&12&457&27&&11&1&1&12&3&&31&2&5&12&1&48\\
\ucca {D}&10&&&&12&280&&40&8&12&2&2&&6&4&1&7&18&20\\
\ucca {E}&48&1&&&20&42&1&294&3&1&17&&&3&7&1&24&4&49\\
\ucca {F}&3&&&&&&&&613&&&&&1&1&&3&&1\\
\ucca {G}&&2&&&&&&&2&6&2&&&&&&2&&4\\
\ucca {H}&40&2&&1&29&6&&13&1&&450&4&&22&&2&8&&265\\
\ucca {L}&&&&&&7&&1&19&1&&221&14&1&&27&&&5\\
\ucca {N}&&&&&1&1&&1&&&&10&31&&1&&&&2\\
\ucca {P}&3&&&&16&15&1&2&13&12&1&1&&345&&2&29&&32\\
\ucca {Q}&&&&&8&5&&1&&&&&&&40&&&&1\\
\ucca {R}&3&&&&6&&&&&&&13&&1&&211&14&&3\\
\ucca {S}&6&&&&48&49&&4&26&&6&&&10&&1&251&&5\\
\ucca {T}&2&&&&4&2&3&&&&&&&&1&&&45&5\\
\ucca {$\emptyset$}&148&1&3&6&136&60&&100&32&1&124&9&2&65&12&34&23&6&
\end{tabularx}
\caption{dev set confusion matrix for the \textbf{rule-based converter}.
The last column (row) %, labeled $\emptyset$,
shows the number of predicted (gold-standard) edges of each category
that do not match any gold-standard (predicted) unit.
\label{tab:confusion_matrix}}
\end{table*}

\section{Analysis}\label{sec:discussion}

\Cref{tab:confusion_matrix} presents the EWT Reviews dev confusion matrix for
the converter's output and gold UCCA.
The delexicalized parsers' confusion matrix (in \cref{sec:parser_confusion}) is similar.\footnote{For multiple UCCA units with the same terminal yield (i.e., units with a single non-remote child),
we take the top category only, to avoid double-counting.}
Note that we consulted the \textit{training} set iteratively while developing the rules, addressing many recurring issues that would show up as prominent confusions.\footnote{A full report
of dev set outputs is included in \url{https://github.com/danielhers/streusle/blob/streusle2ucca/uccareport.dev.tsv}.}

We proceed with an extensive error analysis of the converter,
to point out similarities and delineate remaining
divergences, which we stipulate  constitute content differences between UCCA and the combination of syntax and lexical semantics from UD and STREUSLE.
\Cref{fig:analysis_examples} shows gold annotation and the converter's predictions.

\begin{figure*}[t]
    \centering\scriptsize 
    \renewcommand\arraystretch{1.4}
    \setlength\tabcolsep{4pt}
    \begin{tabular}{llll}
        \normalsize\textbf{STREUSLE Annotation} & \normalsize\textbf{Predicted UCCA Annotation} & \normalsize\textbf{Gold UCCA Annotation} \\
        \hline
        \bf Noun compounds \\
        $\underset{\sst{n.substance}}{\text{tap\_water}}$ & tap water (\textit{unanalyzable}) & [\ucca{E} tap] [\ucca{C} water] & \xmark \\
        $\underset{\sst{n.event}}{\text{road\_construction}}$ & [\ucca{P} road construction] & [\ucca{A} road] [\ucca{P} construction] & \xmark \\
        \bf Adverbs and linkage \\
        $\underset{\sst{v.vid}}{\text{Gets\_busy}}$ so $\underset{\sst{v.motion}}{\text{come}}$ early & [\ucca{H} [\ucca{P} Gets busy] ] [\ucca{L} so] [\ucca{H} [\ucca{P} come] [\ucca{T} early]~] & [\ucca{H} [\ucca{D} Gets] [\ucca{S} busy]~] [\ucca{L} so] [\ucca{H} [\ucca{P} come] [\ucca{T} early]~] & \xcmark
 \\
        so easy to $\underset{\sst{v.motion}}{\text{load}}$ & [\ucca{L} so] [\ucca{H} [\ucca{S} easy] [\ucca{A} [\ucca{F} to] [\ucca{P} load]~]~]  & [\ucca{D} [\ucca{E} so] [\ucca{C} easy]~] [\ucca{F} to] [\ucca{P} load]  & \xmark \\
        \bf Scene-evoking nouns \\
        a $\underset{\sst{n.food}}{\text{meal}}$ $\underset{\sst{p.locus}}{\text{on}}$ the $\underset{\sst{n.communication}}{\text{menu}}$ & [\ucca{F} a] [\ucca{C} meal] [\ucca{E} [\ucca{R} on] [\ucca{F} the] [\ucca{C} menu]~]~] & [\ucca{F} a] [\ucca{C} meal] [\ucca{E} [\ucca{R} on] [\ucca{F} the] [\ucca{C} menu]~]~] & \cmark \\
        $\underset{\sst{v.communication}}{\text{answered}}$ all $\underset{\sst{p.gestalt}}{\underset{\sst{p.originator}}{\text{my}}}$ $\underset{\sst{n.communication}}{\text{questions}}$ & [\ucca{P} answered] [\ucca{A} [\ucca{Q} all] [\ucca{A} my] [\ucca{C} questions]~] & [\ucca{P} answered] [\ucca{A} [\ucca{D} all] [\ucca{A} my] [\ucca{P} questions]~] & \xmark
    \end{tabular}
    \caption{Examples for cases where the rule-based converter produced the correct UCCA annotation due to converging analyses,
    as well as cases where it produced a wrong annotation due to a divergence.
    \label{fig:analysis_examples}}
\end{figure*}

% \daniel{More detailed
% ``human inspection of the results'',
% ``Which of these differences are due to the specific set of rules used vs. the formal differences between the two formalisms?''}
% \nss{we didn't do much error analysis on TUPA output, but I looked at the confusion matrix and it resembles the above confusion matrix in terms of which cells have high values. A diff of the two confusion matrix has 5 cells with $>80$ count of differences: TUPA gets fewer H, E, and A correct than the RB system, and more unmatched A and H. TUPA diagonal adds up to 3809. RB diagonal sums to 4213. How can this be if the overall scores are about the same??}

\subsection{High Match---Converging Analyses}

\paragraph{Participants.}

% A
%      R     P
% 2020 70.6  74.7

% A|P
%      R     P

% 2020 0.0   0.0

% A|S
%      R     P

% 2020 20.0  80.0

% {A, A|P, A|S}
%      R     P

% 2020 69.6  76.5

\ucca{A}s are recovered with high precision and recall.
This is generally expected as most syntactic subjects and objects, as well as some obliques and even clauses, signify scene participants.
Where syntax and semantics diverge, STREUSLE supersenses can rule out unlikely candidates.
The most common sources of missed \ucca{A}s are structural errors, i.e., incorrect scene structures, overly flat units containing more than the referential words, or misinterpreted noun compounds (see \cref{sec:low_match} below).

\paragraph{Function words.}

% F
%      R     P
% 2020 84.9  98.6
As evident in \cref{tab:confusion_matrix}, Function words (\ucca{F}) are
accurately predicted. The distinction between words that
contribute to the semantic meaning and those that do not
is preserved between STREUSLE and UCCA, except for some
cases---mainly infinitive ``to''.

\paragraph{Linkers.}

% L
%      R     P

% 2020 83.7  74.7

Linkers (\ucca{L}) are relatively easy: they are prototypically instantiated by syntactic co- and subordinators. To the extent that these are considered adpositional by STREUSLE, their supersense helps disambiguate between inter-scene linkage and Connectors (\ucca{N}) of non-scenes.
% some \ucca{L}s get mistaken for Rs
% The converter has a tendency to overpredict \ucca{L}s, particularly on gold \ucca{F}s and Relators (\ucca{R}), potentially because it has an explicit rule to generate coordinate structures, including linkages, which gets invoked \textit{after} \ucca{F} and \ucca{R} candidates are identified and thus may overwrite them.\jp{is this too technical and hypothetical? I can't think anymore :/}

\subsection{Partial Match---Inferrable by Combining Syntax and Lexical Semantics}

Time (\ucca{T}) and Quantifier (\ucca{Q}) expressions frequently coincide with certain syntactic categories such as adverbs and prepositions, and can typically be identified from corresponding supersenses, if available.
The converter tends to err on the conservative side, falling back to Adverbials (\ucca{D}) and Elaborators (\ucca{E}) when it cannot find sufficient explicit semantic evidence.

% T
%      R     P

% 2020 60.0  72.6

% Q
%      R     P

% 2020 58.0  72.7

\subsection{Low Match---Divergences or Insufficient Information}\label{sec:low_match}

\paragraph{Noun compound interpretation.}
Lexical composition in noun compounds evokes various forms of event structures,
which are underspecified by  the meaning of the constituent words
\cite{shwartz2019still}.
While often compounding is used for Elaboration, as in
[\ucca{E} tap] [\ucca{C} water],
it is not necessarily always the case. For example, in [\ucca{C} sea] [\ucca{C} bottom] both ``sea" and ``bottom" are Center, since they reflect part-whole relations.
The modifier may also be a Participant in the scene evoked by the head,
as in [\ucca{A} road] [\ucca{P} construction].
This is partially encoded in STREUSLE, as the fact that the MWE ``road construction''
has the \sst{n.event} supersense indicates that it is scene-evoking,
but it still does not reveal the relationship between the constituent words.

\paragraph{Adverbs and linkage.}
While many syntactic adverbs are semantically Linkers (``well'', ``though''),
neither UD nor STREUSLE distinguish them from Adverbials
(``really'', ``possibly'').
Some adverbs, like ``so'', can serve either role (see~\cref{fig:analysis_examples}),
a distinction that is only made in UCCA.

\paragraph{Centers.}

% C
%      R     P
% 2019 59.1  66.7
% 2020 60.4  67.2

\ucca{C} is often unaligned due to the different notions of
multi-word expressions in STREUSLE and UCCA:
``tap water'' is considered a strong MWE in STREUSLE,
but is internally analyzed (with ``water'' being the Center) in UCCA
(see~\cref{fig:analysis_examples}), leading to an unmatched \ucca{C}.

\subsubsection{Scene-evokers}

While the concept of \textit{scenes} is central to UCCA,  correctly identifying scene-evoking words
is one of the more difficult tasks for our converter.
``Scene-ness'' clearly goes beyond syntax (not all verbs evoke scenes and scenes can be evoked by a wide range of POS) and STREUSLE supersenses in isolation are often too coarse to resolve the question whether a given word evokes a scene and, if so, whether it is a Process (\ucca{P}) or a State (\ucca{S}).
The former decision is generally somewhat easier for the converter (Recall of scene-evokers: 71.3\%) than distinguishing between \ucca{P} (Recall: 69.1\%) and \ucca{S} (64.0\%).
Below we examine a few recurring phenomena involving scenes.

% polysemy, context-dependent
% cannot be fully resolved by the intersection of syntax and lexical semantics

% look at P \& R of \{P, S\}. is it much higher than the separate scores?

% P
%      R     P
% 2020 69.1  73.1

% S
%      R     P
% 2020 64.0  61.2

% {P, S}
%      R     P
% 2020 71.3  72.3

% {A|P, A|S, P, S}
%      R     P
% 2020 67.9 72.4

\paragraph{Scene-evoking nouns.}
STREUSLE underspecifies whether nouns evoke scenes.
For example,
``menu'' and ``question'' both have the \sst{n.communication} supersense
in STREUSLE, but ``menu'' does not evoke a scene, while ``question'' might.
Other similarly broad supersenses include \sst{n.cognition} (``decision'': potential scene, ``fact'': non-scene) and \sst{n.possession} (``purchase'': potential scene, ``money'': non-scene).

\paragraph{Relational nouns.}
This is a special case of scene-evoking nouns \cite{newell-18,meyers-04}, both refering to an entity and evoking a scene in which the entity generally or habitually participates.\footnote{E.g., a \textit{teacher} is a person who teaches, and a \textit{friend} is a person in a friendship relation with another person.}
These units have two categories in UCCA, either \ucca{A|P} or \ucca{A|S}.
The converter relies here on a combination of \sst{n.person} or \sst{n.group} supersenses and lexical lists.
However, these nouns' scene-ness is often not recognized and they are confused with regular \ucca{A} or \ucca{C}.
% \jp{is this a bug? see slack}

\paragraph{Scene-evoking adjectives.}
Inspecting the high-frequency confusions, \textbf{adjectives} stand out as persistent error inducers. Different classes of adjectives are handled differently in UCCA: e.g., while most adjectives are scene-evoking, pertainyms (\textit{academic}), inherent-composition modifiers (\textit{sugary}), and quantity modifiers (\textit{many}) are not. Some adjectives are ambiguous: a \textit{legal practice} may refer to a behavior that is legal as opposed to illegal, in which case it should be scene-evoking, or to a law office, in which case it should not. Enriching STREUSLE with supersenses for adjectives \cite{tsvetkov-14} might be fruitful for such distinctions. Even with lexical disambiguation, the scene attachment of the adjective may be ambiguous: e.g.~\textit{a good chef} probably means a chef who cooks well, so \textit{good} should be an Adverbial in the scene evoked by {\it chef}---in contrast with \textit{a tall chef}, where \textit{tall} is not part of the cooking scene and instead should evoke a State.
Predicative adjectives, and adjective modifiers in predicative NPs, are another source of difficulty, especially when they occur in fragments:
sometimes the adjective is annotated as evoking the main scene, and sometimes not. Determining this requires making various semantic distinctions, which are not fully represented in STREUSLE.
% So-called \textit{tough}-adjectives (e.g.,~\textit{\textbf{easy} to load}) involve special syntax that is not currently handled properly, resulting in unmatched units.

% \paragraph{Scene-evoking possessives.}
% STREUSLE underspecifies whether possessives evoke scenes.
% Possession nous, acquisition

% \paragraph{Dates.}
% STREUSLE underspecifies whether a phrase is a date, which is always unanalyzable in uCCA.

% \paragraph{Secondary verbs and existentials.}
% requires lexical lists

% \paragraph{Process vs. State.}
% Coarse grained semantic division that does not differentiate P/S/non-scene

%Finally, coordinated scene-evoking adjectives (\textit{professional and friendly}), which are popular in the genre, receive a special designation to simplify annotation. The data processing needs to be updated to express the implied component scenes.\oa{I think we can leave it out for now; we don't have space, and that's a technical issue we'll fix for CR anyway}

% Error analysis also reveals that \textbf{occupational nouns} (e.g., {\it chef}) are inconsistently annotated in the gold standard UCCA. Sometimes they are not scene-evoking, and sometimes they evoke States, when they should evoke Processes.
%2)~Reflexive pronouns are by UCCA policy part of the predicate, but treated as Participants by the converter. This is easily fixed.\nss{only affects 3 tokens?}

% \jp{add some interesting bits about comparison with 2019 converter}

\section{Conclusion}\label{sec:conclusion}

We have presented an extensive analysis of the similarities and differences between STREUSLE and UCCA on the EWT Reviews corpus,
assisted by two complementary methods:
manual rule-based conversion, and delexicalized parsing.
Both approaches arrived at similar results, showing that the conversion
between the frameworks can be moderately accurate, while also revealing
important divergences, namely distinctions made in UCCA but not in STREUSLE:
semantic relation between nouns in compounds,
adverbial and linkage usage of adverbs,
and the scene-evoking status of nouns, possessives and adjectives,
among others.

Enriching supervised parsers with lexical semantic features improves parsing performance when using gold input. While this paper focuses on analysis, future work will investigate using predicted features with a parser/tagger \cite{liu2020lexical}.
This approach is expected to improve parsing performance and robustness, demonstrating the utility of linguistically-informed approaches in complementing general supervised semantic parsers.

\section*{Acknowledgements}

This research was supported in part by grant 2016375 from the United States--Israel Binational Science Foundation (BSF), Jerusalem, Israel. ML is funded by a Google Focused Research Award. We acknowledge the computational resources provided by CSC in Helsinki and Sigma2 in Oslo through NeIC-NLPL (www.nlpl.eu).

\bibliography{references,mrp}
 \bibliographystyle{acl_natbib}

\newpage

\appendix
\section{Details of Rule-based Converter}\label{sec:rules}

The following is a detailed description of the rules used in the
rule-based parser (\Cref{sec:converter}).
This gives a step-by-step overview of the algorithm for constructing an UCCA semantic graph using STREUSLE/UD annotations.
It is not a full specification and omits many details of the criteria and operations, but should be helpful for understanding the full code, available in \url{https://github.com/danielhers/streusle/blob/streusle2ucca/conllulex2ucca.py}.
A running example is given at each step for the sentence:

\begin{quote}
There's plenty of parking, and I've never had an issue with audience
members who won't stop talking or answering their cellphones.
\end{quote}

It has the following gold annotation:

\begin{quote}
{[}H {[}\textbf{F} There{]} {[}F 's{]} {[}\textbf{D} plenty{]}
{[}\textbf{P} {[}R of{]} {[}C parking{]} {]} {]} {[}U ,{]} {[}L and{]}
{[}H {[}A I{]} {[}F 've{]} {[}\textbf{D\textbar{}T} never{]} {[}F had{]}
{[}P {[}F an{]} {[}C issue{]} {]} {[}A {[}R with{]} {[}C {[}\textbf{A}
audience{]} {[}\textbf{P} members{]} {]} {[}E {[}R who{]} {[}\textbf{H}
{[}A* members{]} {[}F wo{]} {[}D n't{]} {[}D stop{]} {[}P talking{]} {]}
{[}L or{]} {[}\textbf{H} {[}A* members{]} {[}P answering{]} {[}A {[}E
{[}S\textbar{}A their{]} {[}A* cellphones{]} {]} {[}C cellphones{]} {[}U
.{]} {]} {]} {]} {]} {]}
\end{quote}

% \Tree[.H [.\textbf{F} There ] [.F 's ] [.\textbf{D} plenty ] [.\textbf{P} [.R of ] [.C parking ]]]

% \scalebox{.3}{
% \Tree[X [.H [.**F** There ] [.F 's ] [.**D** plenty ] [.**P** [.R of ] [.C parking ]  ]  ] [.U , ] [.L and ] [.H [.A I ] [.F 've ] [.**D|T** never ] [.F had ] [.P [.F an ] [.C issue ]  ] [.A [.R with ] [.C [.**A** audience ] [.**P** members ]  ] [.E [.R who ] [.**H** [.A* members ] [.F wo ] [.D n't ] [.D stop ] [.P talking ]  ] [.L or ] [.**H** [.A* members ] [.P answering ] [.A [.E [.S|A their ] [.A* cellphones ]  ] [.C cellphones ] [.U . ]  ]  ]  ]  ]  ] ]
% }

% [H [**F** There] [F 's] [**D** plenty] [**P** [R of] [C parking] ] ] [U ,] [L and] [H [A I] [F 've] [**D|T** never] [F had] [P [F an] [C issue] ] [A [R with] [C [**A** audience] [**P** members] ] [E [R who] [**H** [A* members] [F wo] [D n't] [D stop] [P talking] ] [L or] [**H** [A* members] [P answering] [A [E [S|A their] [A* cellphones] ] [C cellphones] [U .] ] ] ] ] ]

\subsection*{Step 0.1: Transform the UD dependency parse}

  Split the final preposition off from \lexcat{V.IAV} MWEs like
  \emph{take care of}, as it is usually not treated as part of the
  verbal unit in UCCA. If the remaining part is still an MWE, it is
  labeled \lexcat{V.LVC.full} or \lexcat{V.VPC.full} depending on its
  syntax.

\subsection*{Step 0.2: Initialize lexical units under the UCCA
root}

\begin{itemize}
\item
  Each strong lexical expression (single-word or MWE) in STREUSLE is
  treated as an UCCA unit, with the following exceptions:

  \begin{itemize}
  \item
    An MWE with lexcat \lexcat{V.LVC.cause} is broken into two units:
    \texttt{D} for the light verb modifies the main predicate in a
    \texttt{+} unit.
  \item
    An MWE with lexcat \lexcat{V.LVC.full} is broken into two units:
    \texttt{F} for the light verb modifies the main predicate in a
    \texttt{+} unit.
  \item
    An MWE annotation is discarded if it would lead to a cycle in the
    dependencies such that the highest token of the MWE is dependent on
    a token outside of the MWE, which is dependent on another token of
    the MWE.
  \end{itemize}
\item
  MWE-internal dependency edges are discarded so they will not be
  processed later.
\item
  Punctuation is marked \texttt{U}.
\item
  Mappings between units and dependency nodes are maintained.
\end{itemize}

\subsection*{Step 0.3: Identify which lexical units are main relations
(scene-evoking)}

In a top-down traversal of the dependency parse, visit each word's
lexical unit and decide whether it evokes a state (\texttt{S}), process
(\texttt{P}), undetermined between state or process (\texttt{+}), or
does not evoke a scene (\texttt{-}):

\begin{itemize}
\item
  If already labeled \texttt{D}, \texttt{F}, or \texttt{+}, do nothing.
\item
  If an adjective not from a small list of quantity adjectives, label
  \texttt{S}.
\item
  If existential \emph{there}, label \texttt{S}. If a \emph{be} verb in
  an existential construction, label \texttt{-} and swap the positions
  of \emph{be} and \emph{there} in the dependency parse so \emph{there}
  is the head.
\item
  If an adverb not attached as discourse and it has a copula, label
  \texttt{S}.
\item
  \emph{thanks} and \emph{thank you} are \texttt{P}.
\item
  In most cases, predicative prepositions are \texttt{S}.
\item
  A copula introducing a predicate nominal (non-PP) is labeled
  \texttt{S} and promoted to the head of the dependency parse, unless
  the nominal is scene-evoking. (The top-down traversal order ensures
  the nominal is reached first.)
\item
  If a common noun, mark as

  \begin{itemize}
  \item
    \texttt{S} if supersense-tagged as \sst{attribute}, \sst{feeling}, or \sst{state}
  \item
    \texttt{P} if \sst{act}, \sst{phenomenon}, \sst{process}, or \sst{event} (with the exception
    of nouns denoting a part of the day)
  \item
    a relational noun if \sst{person} or \sst{group} and
    matching kinship/occupation lists or suffixes
  \item
    \texttt{-} otherwise
  \end{itemize}
\item
  If a verb or copula not handled above, label \texttt{+}
\item
  Else label \texttt{-}
\end{itemize}

\emph{In the notation, ``UNA'' means ``lexical'' (it originally meant
``unanalyzable'').}

\begin{quote}
{[}DUMMYROOT {[}S {[}UNA There{]} {]} {[}- {[}UNA 's{]} {]} {[}- {[}UNA
plenty{]} {]} {[}- {[}UNA of{]} {]} {[}S {[}UNA parking{]} {]} {[}U ,{]}
{[}- {[}UNA and{]} {]} {[}- {[}UNA I{]} {]} {[}- {[}UNA 've{]} {]} {[}-
{[}UNA never{]} {]} {[}+ {[}F had{]} \ldots{} {[}UNA issue{]} {]} {[}-
{[}UNA an{]} {]} \ldots{} {[}- {[}UNA with{]} {]} {[}- {[}UNA
audience{]} {]} {[}- {[}UNA\textbar A\textbar P {[}UNA members{]} {]}
{]} {[}- {[}UNA who{]} {]} {[}- {[}UNA wo{]} {]} {[}- {[}UNA n't{]} {]}
{[}+ {[}UNA stop{]} {]} {[}+ {[}UNA talking{]} {]} {[}- {[}UNA or{]} {]}
{[}+ {[}UNA answering{]} {]} {[}- {[}UNA their{]} {]} {[}- {[}UNA
cellphones{]} {]} {[}U .{]} {]}
\end{quote}

\subsection*{Step 1: Attach functional and discourse modifier
words}

Determiners, auxiliaries, copulas are generally \texttt{F}; vocatives
and interjections, \texttt{G}. Exceptions include modal auxiliaries
(\texttt{D}), demonstrative determiners modifying a non-scene unit
(\texttt{E}), quantifier determiners modifying a non-scene unit
(\texttt{Q}).

\emph{Omitting the root:}

\begin{quote}
{[}S {[}UNA There{]} {[}\textbf{F} {[}UNA 's{]} {]} {]} {[}- {[}UNA
plenty{]} {]} {[}- {[}UNA of{]} {]} {[}S {[}UNA parking{]} {]} {[}U ,{]}
{[}- {[}UNA and{]} {]} {[}- {[}UNA I{]} {]} {[}+ {[}\textbf{F} {[}UNA
've{]} {]} \ldots{} {[}\textbf{F} had{]} {[}\textbf{F} {[}UNA an{]} {]}
{[}UNA issue{]} {]} {[}- {[}UNA never{]} {]} \ldots{} {[}- {[}UNA
with{]} {]} {[}- {[}UNA audience{]} {]} {[}- {[}UNA\textbar A\textbar P
{[}UNA members{]} {]} {]} {[}- {[}UNA who{]} {]} {[}+ {[}F {[}UNA wo{]}
{]} \ldots{} {[}UNA stop{]} {]} {[}- {[}UNA n't{]} {]} \ldots{} {[}+
{[}UNA talking{]} {]} {[}- {[}UNA or{]} {]} {[}+ {[}UNA answering{]} {]}
{[}- {[}UNA their{]} {]} {[}- {[}UNA cellphones{]} {]} {[}U .{]}
\end{quote}

\subsection*{Step 2: Attach other modifiers: adverbial,
adjectival, numeric, compound, possessive, predicative-PP, adnominal-PP;
as well as possessive clitic and preposition (as \texttt{R}, unless
possessive clitic marks canonical possession in which case it is
\texttt{S})}

\begin{quote}
{[}S {[}UNA There{]} {[}F {[}UNA 's{]} {]} {]} {[}- {[}UNA plenty{]}
{[}\textbf{E} {[}S {[}\textbf{R} {[}UNA of{]} {]} {[}UNA parking{]} {]}
{]} {]} {[}U ,{]} {[}- {[}UNA and{]} {]} {[}- {[}UNA I{]} {]} {[}+ {[}F
{[}UNA 've{]} {]} {[}\textbf{T} {[}UNA never{]} {]} {[}F had{]} {[}F
{[}UNA an{]} {]} {[}UNA issue{]} {[}A {[}\textbf{R} {[}UNA with{]} {]}
{[}\textbf{E} {[}UNA audience{]} {]} {[}UNA\textbar{}A\textbar{}P {[}UNA
members{]} {]} \ldots{} {[}\textbf{E} {[}+ \textbf{{[}A* members{]}}
{[}\textbf{F} {[}UNA wo{]} {]} {[}\textbf{D} {[}UNA n't{]} {]} {[}UNA
stop{]} {]} {]} {]} {]} {[}- {[}UNA who{]} {]} \ldots{} {[}+ {[}UNA
talking{]} {]} {[}- {[}UNA or{]} {]} {[}+ {[}UNA answering{]} {]} {[}-
{[}\textbf{E} {[}\textbf{A\textbar{}S} {[}UNA their{]} {]} \textbf{{[}A*
cellphones{]}} {]} {[}UNA cellphones{]} {]} {[}U .{]}
\end{quote}

\subsection*{Step 3: Process verbal argument structure relations:
subjects, objects, obliques, clausal complements; flag secondary
(non-auxiliary) verb
constructions}

\begin{quote}
{[}S {[}UNA There{]} {[}F {[}UNA 's{]} {]} {[}\textbf{A} {[}UNA
plenty{]} {[}E {[}S {[}R {[}UNA of{]} {]} {[}UNA parking{]} {]} {]} {]}
{]} {[}U ,{]} {[}- {[}UNA and{]} {]} {[}+ {[}\textbf{A} {[}UNA I{]} {]}
{[}F {[}UNA 've{]} {]} {[}T {[}UNA never{]} {]} {[}F had{]} {[}F {[}UNA
an{]} {]} {[}UNA issue{]} {[}A {[}R {[}UNA with{]} {]} {[}E {[}UNA
audience{]} {]} {[}UNA\textbar{}A\textbar{}P {[}UNA members{]} {]} {[}E
{[}+ {[}A* members{]} {[}R {[}UNA who{]} {]} {[}F {[}UNA wo{]} {]} {[}D
{[}UNA n't{]} {]} {[}UNA stop{]} {[}\textbf{\^{}} {[}+ {[}UNA talking{]}
{]} {]} {]} {]} {]} {]} {[}- {[}UNA or{]} {]} {[}+ {[}UNA answering{]}
{[}\textbf{A} {[}E {[}A\textbar{}S {[}UNA their{]} {]} {[}A*
cellphones{]} {]} {[}UNA cellphones{]} {]} {]} {[}U .{]}
\end{quote}

\subsection*{Step 4: Coordination}

Traversing the graph top-down: for each coordinate construction with
conjuncts' units' categories X and Y, create a ternary-branching
structure \texttt{{[}X(COORD)\ X\ L\ Y{]}} if X is scene-evoking
(\texttt{+}, \texttt{P}, or \texttt{S}) and
\texttt{{[}X(COORD)\ X\ N\ Y{]}} otherwise.

\begin{quote}
There's plenty\ldots{}and I've never had an issue with\ldots{}
\end{quote}

\begin{quote}
{[}\textbf{S(COORD)} {[}S {[}UNA There{]} {[}F {[}UNA 's{]} {]} {[}A
{[}UNA plenty{]} {[}E {[}S {[}R {[}UNA of{]} {]} {[}UNA parking{]} {]}
{]} {]} {]} \ldots{} {[}\textbf{L} {[}UNA and{]} {]} {[}+ {[}A {[}UNA
I{]} {]} {[}F {[}UNA 've{]} {]} {[}T {[}UNA never{]} {]} {[}F had{]}
{[}F {[}UNA an{]} {]} {[}UNA issue{]} {[}A {[}R {[}UNA with{]} {]} {[}E
{[}UNA audience{]} {]} {[}UNA\textbar{}A\textbar{}P {[}UNA members{]}
{]} {[}E {[}+ {[}A* members{]} {[}R {[}UNA who{]} {]} {[}F {[}UNA wo{]}
{]} {[}D {[}UNA n't{]} {]} {[}UNA stop{]} {[}\^{} {[}+ {[}UNA talking{]}
{]} {]} {]} {]} {]} {]} {]} {[}U ,{]} \ldots{} {[}- {[}UNA or{]} {]}
{[}+ {[}UNA answering{]} {[}A {[}E {[}A\textbar{}S {[}UNA their{]} {]}
{[}A* cellphones{]} {]} {[}UNA cellphones{]} {]} {]} {[}U .{]}
\end{quote}

\begin{quote}
\ldots{}won't stop talking or answering\ldots{}
\end{quote}

\begin{quote}
{[}S(COORD) {[}S {[}UNA There{]} {[}F {[}UNA 's{]} {]} {[}A {[}UNA
plenty{]} {[}E {[}S {[}R {[}UNA of{]} {]} {[}UNA parking{]} {]} {]} {]}
{]} \ldots{} {[}L {[}UNA and{]} {]} {[}\textbf{P} {[}A {[}UNA I{]} {]}
{[}F {[}UNA 've{]} {]} {[}T {[}UNA never{]} {]} {[}F had{]} {[}F {[}UNA
an{]} {]} {[}UNA issue{]} {[}A {[}R {[}UNA with{]} {]} {[}E {[}UNA
audience{]} {]} {[}UNA\textbar{}A\textbar{}P {[}UNA members{]} {]} {[}E
{[}\textbf{P} {[}A* members{]} {[}R {[}UNA who{]} {]} {[}F {[}UNA wo{]}
{]} {[}D {[}UNA n't{]} {]} {[}UNA stop{]} {[}\^{} {[}+(COORD)
{[}\textbf{P} {[}UNA talking{]} {]} {[}L {[}UNA or{]} {]} {[}\textbf{P}
{[}UNA answering{]} {[}A {[}E {[}A\textbar{}S {[}UNA their{]} {]} {[}A*
cellphones{]} {]} {[}UNA cellphones{]} {]} {]} {]} {]}{]} {]} {]} {]}
{]} {[}U ,{]} \ldots{} {[}U .{]}
\end{quote}

\subsection*{Step 5: Decide \texttt{S} or \texttt{P} for
remaining \texttt{+}
scenes}

Copula \emph{be} and stative \emph{have} are \texttt{S}; other verbs, as
well as nouns tagged as ACT, PHENOMENON, PROCESS, or EVENT, are
\texttt{P}.

\begin{quote}
{[}S(COORD) {[}S {[}UNA There{]} {[}F {[}UNA 's{]} {]} {[}A {[}UNA
plenty{]} {[}E {[}S {[}R {[}UNA of{]} {]} {[}UNA parking{]} {]} {]} {]}
{]} \ldots{} {[}L {[}UNA and{]} {]} {[}\textbf{P} {[}A {[}UNA I{]} {]}
{[}F {[}UNA 've{]} {]} {[}T {[}UNA never{]} {]} {[}F had{]} {[}F {[}UNA
an{]} {]} {[}UNA issue{]} {[}A {[}R {[}UNA with{]} {]} {[}E {[}UNA
audience{]} {]} {[}UNA\textbar{}A\textbar{}P {[}UNA members{]} {]} {[}E
{[}\textbf{P} {[}A* members{]} {[}R {[}UNA who{]} {]} {[}F {[}UNA wo{]}
{]} {[}D {[}UNA n't{]} {]} {[}UNA stop{]} {[}\^{} {[}+(COORD)
{[}\textbf{P} {[}UNA talking{]} {]} {[}L {[}UNA or{]} {]} {[}\textbf{P}
{[}UNA answering{]} {[}A {[}E {[}A\textbar{}S {[}UNA their{]} {]} {[}A*
cellphones{]} {]} {[}UNA cellphones{]} {]} {]} {]} {]}{]} {]} {]} {]}
{]} {[}U ,{]} \ldots{} {[}U .{]}
\end{quote}

\subsection*{Step 6.1: Restructure for secondary verbs}

\begin{quote}
{[}S(COORD) {[}S {[}UNA There{]} {[}F {[}UNA 's{]} {]} {[}A {[}UNA
plenty{]} {[}E {[}S {[}R {[}UNA of{]} {]} {[}UNA parking{]} {]} {]} {]}
{]} \ldots{} {[}L {[}UNA and{]} {]} {[}P {[}A {[}UNA I{]} {]} {[}F
{[}UNA 've{]} {]} {[}T {[}UNA never{]} {]} {[}F had{]} {[}F {[}UNA an{]}
{]} {[}UNA issue{]} {[}A {[}R {[}UNA with{]} {]} {[}E {[}UNA audience{]}
{]} {[}UNA\textbar{}A\textbar{}P {[}UNA members{]} {]} {[}E {[}P {[}A*
members{]} {[}R {[}UNA who{]} {]} {[}F {[}UNA wo{]} {]} {[}D {[}UNA
n't{]} {]} {[}\textbf{D} {[}UNA stop{]} {]} {[}\textbf{+(COORD)} {[}P
{[}UNA talking{]} {]} {[}L {[}UNA or{]} {]} {[}P {[}UNA answering{]}
{[}A {[}E {[}A\textbar{}S {[}UNA their{]} {]} {[}A* cellphones{]} {]}
{[}UNA cellphones{]} {]} {]} {]}{]} {]} {]} {]} {]} {[}U ,{]} \ldots{}
{[}U .{]}
\end{quote}

\subsection*{Step 6.2: Articulation---marking lexical
heads of units as \texttt{C}, \texttt{P}, or \texttt{S}, and renaming
scene units as \texttt{H} where necessary; determination of \texttt{C}
involves ``X of Y'' constructions involving
quantities/Species}

\begin{quote}
{[}S(COORD) {[}\textbf{H(S)} {[}S {[}UNA There{]} {]} {[}F {[}UNA 's{]}
{]} {[}A {[}\textbf{Q} {[}UNA plenty{]} {]} {[}E {[}\textbf{H(S)} {[}R
{[}UNA of{]} {]} {[}\textbf{S} {[}UNA parking{]} {]} {]} {]} {]} {]}
\ldots{} {[}L {[}UNA and{]} {]} {[}\textbf{H(P)} {[}A {[}UNA I{]} {]}
{[}F {[}UNA 've{]} {]} {[}T {[}UNA never{]} {]} {[}F had{]} {[}F {[}UNA
an{]} {]} {[}\textbf{P} {[}UNA issue{]} {]} {[}A {[}R {[}UNA with{]} {]}
{[}E {[}UNA audience{]} {]}
{[}\textbf{H(A\textbar{}P)\textbar{}A\textbar{}P}
{[}\textbf{A\textbar{}P} {[}UNA members{]} {]} {]} {[}E {[}\textbf{H}
{[}A* members{]} {[}R {[}UNA who{]} {]} {[}F {[}UNA wo{]} {]} {[}D
{[}UNA n't{]} {]} {[}D {[}UNA stop{]} {]} {[}+(COORD) {[}\textbf{H(P)}
{[}P {[}UNA talking{]} {]} {]} {[}L {[}UNA or{]} {]} {[}\textbf{H(P)}
{[}P {[}UNA answering{]} {]} {[}A {[}E
{[}\textbf{H(A\textbar{}S)\textbar{}S} {[}A\textbar{}S {[}UNA their{]}
{]} {]} {[}A* cellphones{]} {]} {[}\textbf{C} {[}UNA cellphones{]} {]}
{]} {]} {]} {]} {]} {]} {]} {]} {[}U ,{]} \ldots{} {[}U .{]}
\end{quote}

\subsection*{Steps 7+: Cleanup}

Remove temporary decorations on \texttt{H} units from articulation; move
\texttt{U} units for punctuation to more convenient attachment points;
convert remaining \texttt{-} and \texttt{+} labels; wrap stray
\texttt{P} and \texttt{S} units with \texttt{H} scenes; remove
\texttt{UNA} and other temporary designations in the graph

\begin{quote}
{[}S(COORD) {[}\textbf{H} {[}S {[}UNA There{]} {]} {[}F {[}UNA 's{]} {]}
{[}A {[}Q {[}UNA plenty{]} {]} {[}E {[}\textbf{H} {[}R {[}UNA of{]} {]}
{[}S {[}UNA parking{]} {]} {]} {]} {]} {]} \ldots{} {[}L {[}UNA and{]}
{]} {[}\textbf{H} {[}A {[}UNA I{]} {]} {[}F {[}UNA 've{]} {]} {[}T
{[}UNA never{]} {]} {[}F had{]} {[}F {[}UNA an{]} {]} {[}P {[}UNA
issue{]} {]} {[}A {[}R {[}UNA with{]} {]} {[}E {[}UNA audience{]} {]}
{[}\textbf{C} {[}A\textbar{}P {[}UNA members{]} {]} {]} {[}E {[}H {[}A*
members{]} {[}R {[}UNA who{]} {]} {[}F {[}UNA wo{]} {]} {[}D {[}UNA
n't{]} {]} {[}D {[}UNA stop{]} {]} {[}+(COORD) {[}\textbf{H} {[}P {[}UNA
talking{]} {]} {]} {[}L {[}UNA or{]} {]} {[}\textbf{H} {[}P {[}UNA
answering{]} {]} {[}A {[}E {[}\textbf{C} {[}A\textbar{}S {[}UNA their{]}
{]} {]} {[}A* cellphones{]} {]} {[}C {[}UNA cellphones{]} {]} {]} {]}
{]} {]} {]} {]} {]} {]} {[}U ,{]} \ldots{} {[}U .{]}
\end{quote}

\begin{quote}
{[}S(COORD) {[}H {[}S {[}UNA There{]} {]} {[}F {[}UNA 's{]} {]} {[}A
{[}Q {[}UNA plenty{]} {]} {[}E {[}H {[}R {[}UNA of{]} {]} {[}S {[}UNA
parking{]} {]} {]} {]} {]} {]} \textbf{{[}U ,{]}} {[}L {[}UNA and{]} {]}
{[}H {[}A {[}UNA I{]} {]} {[}F {[}UNA 've{]} {]} {[}T {[}UNA never{]}
{]} {[}F had{]} {[}F {[}UNA an{]} {]} {[}P {[}UNA issue{]} {]} {[}A {[}R
{[}UNA with{]} {]} {[}E {[}UNA audience{]} {]} {[}C {[}A\textbar{}P
{[}UNA members{]} {]} {]} {[}E {[}H {[}A* members{]} {[}R {[}UNA who{]}
{]} {[}F {[}UNA wo{]} {]} {[}D {[}UNA n't{]} {]} {[}D {[}UNA stop{]} {]}
{[}+(COORD) {[}H {[}P {[}UNA talking{]} {]} {]} {[}L {[}UNA or{]} {]}
{[}H {[}P {[}UNA answering{]} {]} {[}A {[}E {[}C {[}A\textbar{}S {[}UNA
their{]} {]} {]} {[}A* cellphones{]} {]} {[}C {[}UNA cellphones{]} {]}
{]} {]} {]} {]} {]} {]} {]} \textbf{{[}U .{]}} {]}
\end{quote}

\begin{quote}
{[}\textbf{H} {[}H {[}S {[}UNA There{]} {]} {[}F {[}UNA 's{]} {]} {[}A
{[}Q {[}UNA plenty{]} {]} {[}E {[}H {[}R {[}UNA of{]} {]} {[}S {[}UNA
parking{]} {]} {]} {]} {]} {]} {[}U ,{]} {[}L {[}UNA and{]} {]} {[}H
{[}A {[}UNA I{]} {]} {[}F {[}UNA 've{]} {]} {[}T {[}UNA never{]} {]}
{[}F had{]} {[}F {[}UNA an{]} {]} {[}P {[}UNA issue{]} {]} {[}A {[}R
{[}UNA with{]} {]} {[}E {[}UNA audience{]} {]} {[}C {[}A\textbar{}P
{[}UNA members{]} {]} {]} {[}E {[}H {[}A* members{]} {[}R {[}UNA who{]}
{]} {[}F {[}UNA wo{]} {]} {[}D {[}UNA n't{]} {]} {[}D {[}UNA stop{]} {]}
{[}\textbf{H} {[}H {[}P {[}UNA talking{]} {]} {]} {[}L {[}UNA or{]} {]}
{[}H {[}P {[}UNA answering{]} {]} {[}A {[}E {[}C {[}A\textbar{}S {[}UNA
their{]} {]} {]} {[}A* cellphones{]} {]} {[}C {[}UNA cellphones{]} {]}
{]} {]} {]} {]} {]} {]} {]} {[}U .{]} {]}
\end{quote}

\begin{quote}
{[}H {[}H {[}S There{]} {[}F 's{]} {[}A {[}Q plenty{]} {[}E {[}R of{]}
{[}S parking{]} {]} {]} {]} {[}U ,{]} {[}L and{]} {[}H {[}A I{]} {[}F
've{]} {[}T never{]} {[}F had{]} {[}F an{]} {[}P issue{]} {[}A {[}R
with{]} {[}E audience{]} {[}C {[}A\textbar{}P members{]} {]} {[}E {[}A*
members{]} {[}R who{]} {[}F wo{]} {[}D n't{]} {[}D stop{]} {[}H {[}H
{[}P talking{]} {]} {[}L or{]} {[}H {[}P answering{]} {[}A {[}E {[}C
{[}A\textbar{}S their{]} {]} {[}A* cellphones{]} {]} {[}C cellphones{]}
{]} {]} {]} {]} {]} {]} {[}U .{]} {]}
\end{quote}

\section{An Alternative Converter}\label{sec:altconverter}

Prior to the converter described in this paper, we had developed a converter based on
the UD-to-UCCA converter of \newcite{hershcovich2019content}.
However, after rewriting the converter from scratch in a more modular fashion,
yielding the current converter (\S\ref{sec:converter}), we found that it obtains better performance, and therefore
abandoned the previous, alternative converter.
For completeness, we describe the previous converter here. The source code is available online.\footnote{\url{https://github.com/danielhers/streusle/blob/streusle2ucca-2019/conllulex2ucca.py}}

\subsection{Dependency Transformations}
\newcite{hershcovich2019content}
apply several pre-conversion dependency transformation,
to match UCCA's flat linkage structure:
\texttt{cc} dependents and \texttt{mark} dependents of \texttt{advcl}
are promoted to be siblings of their heads, as in UCCA they are often \ucca{L}
between scenes.
\texttt{advcl}, \texttt{appos}, \texttt{conj} and \texttt{parataxis}
are also promoted. We implement the same transformations using
DepEdit.\footnote{\url{https://corpling.uis.georgetown.edu/depedit}}

\subsection{Label Mapping}\label{sec:mapping}
We assign UCCA categories to graph edges,
translating each UD relation to the
most commonly co-occurring UCCA category in the training set.

We augment this simple majority-based UD-to-UCCA category mapping
with rules based on lexical semantics, whose synopsis follows.
The rules are based on the concept of scene-evoking words and phrases, which
is not directly encoded in either UD or STREUSLE.
However, by defining a set of heuristics on the lexical semantic categories,
we obtain a good approximation of the UCCA scene/non-scene
distinction.
These rules first apply %easy (=high confidence according to training corpus statistics) 
heuristics based on STREUSLE noun and verb supersenses---determined by examining the training data---and then take into account POS tags and word lists for the more complex cases.

\paragraph{Scene-evoking or not?}
If a noun has one of the supersenses \sst{n.act}, \sst{n.event}, \sst{n.phenomenon}, \sst{n.process}, \sst{n.state}, \sst{n.attribute}, or \sst{n.feeling}, it is identified as scene-evoking.
Proper nouns (UPOS=PROPN) are classified as non-scene-evoking. 
Other nouns with the supersense \sst{n.person} are first matched against a hand-crafted list of relational suffixes\footnote{\textit{-er}, \textit{-ess}, \textit{-or}, \textit{-ant}, \textit{-ent}, \textit{-ee}, \textit{-ian}, \textit{-ist}} and then against a list of relational nouns based on AMR heuristic lists and WordNet synsets.\footnote{From \url{http://amr.isi.edu/download.html} we obtained \texttt{have-org-role-91-roles-v1.06}, 
which lists 39 types of government officials; and
\texttt{have-rel-role-91-roles-v1.06},
which includes 85 kinship terms and a handful of person-to-person relations like \textit{boss}, \textit{client}, and \textit{roommate}
(the `MAYBE' entries in these lists, which contain ambiguous words, are disregarded).
The WordNet list consists of 1,487 occupations given by the single-word lemmas in the synsets \texttt{leader.n.01}, \texttt{professional.n.01}, \texttt{worker.n.01}, and their hyponyms, minus the words \textit{man} and \textit{woman}.
The heuristics assign the State category to kinship terms like `father' and
Process to occupation nouns.}
%to determine whether they are scene-evoking (if they match at least once) or not.
Copulas with UPOS=AUX and supersense \sst{v.stative}
are classified as non--scene-evoking.
\sst{v.change} verbs are matched against a hand-crafted list of aspectual verbs,\footnote{\textit{start}, \textit{stop}, \textit{begin}, \textit{end}, \textit{finish}, \textit{complete}, \textit{continue}, \textit{resume}, \textit{get}, \textit{become}, \textit{quit}, \textit{keep}} which are non--scene-evoking (\ucca{D}) in UCCA.
Nouns and verbs that do not satisfy any of these criteria are canonically classified as non--scene-evoking and scene-evoking, respectively.

\paragraph{Categories.}
Scene-evoking verbs and nouns are labeled \ucca{P}
instead of \ucca{C} by default
(and their modifiers \ucca{D} instead of \ucca{E}).
However, they are \ucca{S} under the following conditions:
nouns with the supersenses \sst{n.state}, \sst{n.attribute}, and \sst{n.feeling}, and adjectives modifying non-scene-evoking nouns.
Predicative nouns with no subject are labeled as \ucca{A},
  and their modifiers as \ucca{S} (e.g.~``Great food!'').
Possessives with \sst{p.SocialRel} or \sst{p.OrgRole} supersenses
  are \ucca{E} scenes with the possessive labeled
  both \ucca{S} and \ucca{A}, and their head as a \ucca{A} in the scene they evoke (see example in \Cref{fig:ex}).

\paragraph{Special cases of verbs.}
Verbs with the \sst{v.stative} supersense are labeled \ucca{F} if they have
  scene-evoking objects (e.g.,~``they \textit{have} great customer service''),
  as \ucca{S} if their lemma is ``be''\slash ``have'', and as \ucca{P} otherwise.
Verbs in full light verb constructions or verb idiomatic expressions
  (\lexcat{V.LVC.full}, \lexcat{V.VID}) are labeled \ucca{F}
  (e.g.,~``\textit{pay} attention'').
%Causal light verbs (\lexcat{V.LVC.cause}) are labeled as
%  \ucca{D}, as they are in fact secondary relations.

\paragraph{Further lexical decisions.}
Expressions with the \sst{n.time} supersense are labeled \ucca{T}.
Locative pro-adverbs from a pre-defined lexicon (``here'', ``there'', etc.)\ 
  are labeled as \ucca{A}.
Expressions with the \lexcat{NUM} lexcat are labeled \ucca{Q}.
If they are not labeled as \ucca{R} or \ucca{D} by the majority-based mapping, words from a pre-defined lexicon or having the \lexcat{DISC} lexcat or the \sst{p.purpose} supersense are labeled \ucca{L}.
Adverbs with the \sst{p.approximator} supersense are \ucca{E} instead of \ucca{D} (e.g.~``about 30\%'').

\paragraph{Further structural decisions.} 
While \texttt{conj} (conjunct, e.g., in coordination) most often correspond to
\ucca{H}, if the head is a non-scene then the dependent is labeled as \ucca{C} instead, as its corresponding unit is likely non-scene too.
Compound or possessive modifiers of scene nouns are labeled as \ucca{A}.
\texttt{vocative} dependents are labeled as both \ucca{A} and \ucca{G}.

\subsection{UCCA Postprocessing}

UCCA enforces a number of well-formedness restrictions, in terms of which categories may be siblings or children of which.
These are sometimes violated by applying the rules described so far. %are quite accurate (see \cref{sec:results}),
%the resulting graphs still have some inconsistencies that can be resolved
%by looking at the UCCA categories only. 
We apply the category replacements listed in \Cref{tab:postprocessing} to enforce meeting them.
Additionally, \ucca{H} or \ucca{L} inside a scene are promoted to be a sibling
of the scene,
and remote \ucca{H}, \ucca{N}, \ucca{L} are removed.

\begin{table}[ht]
\centering
\begin{tabular}{@{}lcl@{}}
\ucca{C} of unit with \ucca{A} &$\to$& \ucca{P}\\
\ucca{P}, \ucca{S} or \ucca{D} in unit with \ucca{C} and \ucca{N} &$\to$&
  \ucca{C}\\
\ucca{P}, \ucca{S} or \ucca{D} in unit with \ucca{C} and without \ucca{N} &$\to$&
  \ucca{E}\\
\ucca{N} in unit with \ucca{H} &$\to$& \ucca{L}\\
\ucca{L} in unit without \ucca{H}, starting a scene &$\to$& \ucca{R}\\
\ucca{L} in unit without \ucca{H}, not starting a scene &$\to$& \ucca{N}\\
Top-level, category not in
\{\ucca{L},\ucca{H},\ucca{F},\ucca{G},\ucca{U}\} &$\to$& \ucca{H}
\end{tabular}
\caption{UCCA postprocessing category replacements in the alternative converter.\label{tab:postprocessing}}
\end{table}

\subsection{Comparison with the Primary Converter}

% 2019

% Regular edges:
% Precision: 0.696 (4346/6243)
% Recall: 0.702 (4346/6190)
% F1: 0.699

% 2020

% Regular edges:
% Precision: 0.714 (4476/6272)
% Recall: 0.721 (4476/6208)
% F1: 0.717

% A
%      R     P
% 2019 74.4  72.3
% 2020 70.6  74.7

% C
%      R     P
% 2019 59.1  66.7
% 2020 60.4  67.2

% D
%      R     P
% 2019 45.5  56.9
% 2020 35.6  66.4

% E
%      R     P
% 2019 60.1  52.5  (calls UNM, D, A, S, C, H   E; calls E   UNM, S, D, F, A E)
% 2020 61.8  57.1  (calls UNM, A, C, D, S      E; calls E   UNM, DUMMY, C, D)

% F
%      R     P
% 2019 89.9  82.5
% 2020 84.9  98.6

In the following head-to-head comparison, we refer to the primary system presented in the main part of the paper as ``system A'' and the secondary version presented here as ``system B''.

\paragraph{Easy for both:}

\begin{itemize}
    \item both systems perform well on \ucca{A}s
    \item both systems are good at recalling \ucca{F}s (system A: 84.9, system B: 89.9), but system A (in contrast to system B) has almost perfect precision (98.6 vs 82.5)
\end{itemize}

\paragraph{Difficult for both:}

\begin{itemize}
    \item both systems perform okay on \ucca{C}s; system B tends to confuse \ucca{C}s for \ucca{A}s and \ucca{P}s more than system A, which tends to fail at predicting units matching gold \ucca{C}s entirely
    \item \ucca{D}s are difficult for both systems; system A underpredicts \ucca{D}s more than system B, but it is also more precise
    \item \ucca{E}s are difficult for both systems; \ucca{E}s often get confused (by both systems and in both directions) with \ucca{D}s, \ucca{A}s and \ucca{S}s
    \item relational nouns (\ucca{A|S}, \ucca{A|P}) are very difficult for both systems; system B doesn't predict them at all, and system A predicts a few \ucca{A|S}s which are mostly correct, but still misses 3/4 of them (and all \ucca{A|P}s)
    \item \ucca{G}s are very difficult for both systems; system B doesn't predict them at all, and system A predicts a few but with low precision and recall
\end{itemize}

\paragraph{Differences:}

\begin{itemize}
    \item system A is better at recalling \ucca{Q}s
    \item system A is better at recall (64.0 vs 53.6) and precision (61.2 vs 48.4) on \ucca{S}s, but also confuses some gold \ucca{F}s for \ucca{S}s
    \item system A is better at recalling \ucca{T}s; both systems tend to confuse \ucca{T}s for \ucca{D}s, but system B does it more than half of the time whereas system A only a quarter
    \item system A is more eager to predict \ucca{L}s, thus has higher recall (83.7 vs 61.0) but lower precision (74.7 vs 87.5) here than system B; system A confuses some gold \ucca{F}s and \ucca{R}s for \ucca{L}s, system B confuses some gold \ucca{L}s for \ucca{C}s, \ucca{N}s and \ucca{R}s
    \item system B is more eager to predict \ucca{N}s, thus has higher recall (76.6 vs 66.0) but lower precision (37.1 vs 66.0) here than system A; system B confuses some gold \ucca{L}s for \ucca{N}s
    \item system B is more eager to predict \ucca{P}s, thus has higher recall (78.6 vs 69.1) but lower precision (56.3 vs 73.1) here than system A; system B confuses some gold \ucca{C}s, \ucca{D}s and \ucca{F}s for \ucca{P}s, system A confuses some gold \ucca{P}s for \ucca{H}s
    \item system B is more eager to predict \ucca{R}s, thus has higher recall (88.8 vs 74.0) but lower precision (65.5 vs 84.1) here than system A; system B confuses some gold \ucca{F}s and \ucca{L}s for \ucca{R}s, system A confuses some gold \ucca{R}s for \ucca{L}s
\end{itemize}
% }

Given the small differences between the converters in terms of performance, we decided to use system A for the main analysis in the paper, as it is more modular and interpretable.

\section{Delexicalized Parser Confusion Matrix}\label{sec:parser_confusion}

\Cref{tab:parser_confusion_matrix} shows the confusion matrix for the
delexicalized HIT-SCIR parser on the EWT reviews development set.

\begin{table*}[bh]
\centering
\setlength\tabcolsep{4.5pt}
\begin{tabularx}{\textwidth}{@{}lRRRRRRRRRRRRRRRRRRR@{}}
\multicolumn{5}{l}{Predicted Category} & \multicolumn{15}{c}{Gold Category} \\
&\multicolumn{1}{c}{\ucca {A}}&\multicolumn{1}{c}{\ucca {A|G}}&\multicolumn{1}{c}{\ucca {A|P}}&\multicolumn{1}{c}{\ucca {A|S}}&\multicolumn{1}{c}{\ucca {C}}&\multicolumn{1}{c}{\ucca {D}}&\multicolumn{1}{c}{\ucca {D|T}}&\multicolumn{1}{c}{\ucca {E}}&\multicolumn{1}{c}{\ucca {F}}&\multicolumn{1}{c}{\ucca {G}}&\multicolumn{1}{c}{\ucca {H}}&\multicolumn{1}{c}{\ucca {L}}&\multicolumn{1}{c}{\ucca {N}}&\multicolumn{1}{c}{\ucca {P}}&\multicolumn{1}{c}{\ucca {Q}}&\multicolumn{1}{c}{\ucca {R}}&\multicolumn{1}{c}{\ucca {S}}&\multicolumn{1}{c}{\ucca {T}}&\multicolumn{1}{c}{\ucca {$\emptyset$}}\\
\ucca {A}&741&8&3&14&52&14&&40&3&1&15&5&&16&1&5&6&4&199\\ 
\ucca {A|C}&14&1&&2&77&&&1&&1&&&&4&1&&1&&6\\ 
\ucca {A|P}&&&4&1&1&&&&&&&&&2&&&&&\\ 
\ucca {A|Q}&&&&&&&&&&&1&&&&&&&&\\ 
\ucca {A|S}&1&&&6&6&&&&&&&&&1&&&&&1\\ 
\ucca {C}&31&&5&4&406&5&&17&4&1&18&&1&39&2&1&23&1&69\\ 
\ucca {D}&18&&&&18&275&2&43&15&8&7&6&&5&2&1&28&24&25\\ 
\ucca {D|S}&&&&&&&&&&&&&&&&&1&&\\ 
\ucca {D|T}&&&&&1&2&2&&&&&&&&&&&6&\\ 
\ucca {E}&49&&2&&24&36&&169&20&1&9&3&&3&8&2&9&4&147\\ 
\ucca {F}&2&&&&3&42&&16&626&&1&2&&7&12&6&19&&13\\ 
\ucca {G}&&&&&1&&&2&&1&1&1&&&1&&&&\\ 
\ucca {H}&11&1&&&10&&&6&1&1&578&1&&4&&&3&&264\\ 
\ucca {H|R}&&&&&&&&&&&&&&&&&&&1\\ 
\ucca {L}&3&&&&5&6&1&2&10&6&1&187&16&&1&16&2&5&6\\ 
\ucca {N}&&&&&&1&&&&&&14&29&&&5&1&&1\\ 
\ucca {P}&12&&1&5&37&20&&9&10&5&13&1&&357&2&1&53&&43\\ 
\ucca {Q}&&&&&6&5&&4&&&&&&&31&&&&1\\ 
\ucca {R}&10&&&&8&1&&1&15&&&34&&1&&225&11&&12\\ 
\ucca {S}&7&&1&&33&37&&67&10&5&20&&&17&1&4&224&&14\\ 
\ucca {T}&1&&&&1&6&&1&&1&1&2&&&&&1&21&7\\ 
\ucca {$\emptyset$}&173&&1&7&65&55&&98&8&4&273&8&1&38&7&19&11&9
\end{tabularx}
\caption{Development set confusion matrix for the \textbf{delexicalized HIT-SCIR parser}.
The last column (row), labeled $\emptyset$,
shows the number of predicted (gold-standard) edges of each category
that do not match any gold-standard (predicted) unit.
\label{tab:parser_confusion_matrix}}
\end{table*}

\end{document}